\documentclass[10pt,twocolumn,letterpaper]{article}
\usepackage[pagenumbers]{cvpr}
\usepackage{multirow}
\usepackage{amsfonts}
\usepackage{array}
\usepackage{microtype}
\usepackage{placeins}
\usepackage{dblfloatfix}
\usepackage[percent]{overpic}
\definecolor{cvprblue}{rgb}{0.21,0.49,0.74}
\usepackage[pagebackref,breaklinks,colorlinks,allcolors=cvprblue]{hyperref}

\newcommand{\rev}[1]{{\color{black}#1}}

\title{OrthoLatentCAD: Orthographic Spatial Latents for Hand-Drawn CAD Program Generation}
\author{Xiaolei Zhou \quad Xinjie Zhu \quad Chuangjie Fang \\ Chenyu Wang \quad Jianwei Zheng \\ Zhejiang University of Technology}

\begin{document}
\twocolumn[{%
\renewcommand\twocolumn[1][]{#1}%
\maketitle
\begin{minipage}{\textwidth}
    \centering
    \includegraphics[width=0.85\textwidth]{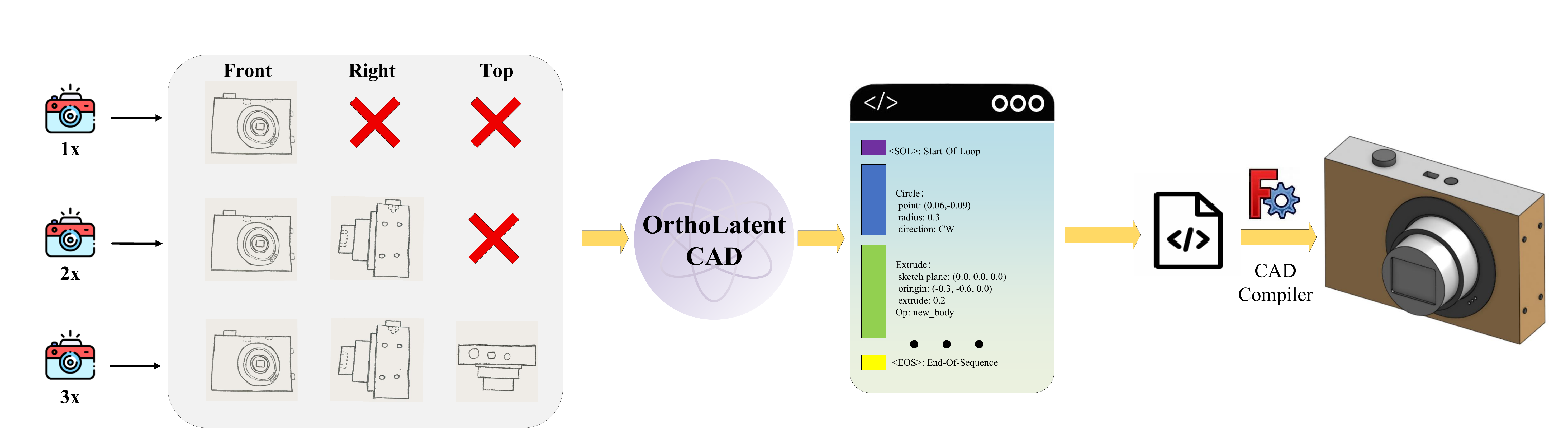}
    \captionsetup{hypcap=false}
    \captionof{figure}{\textbf{Task overview.} OrthoLatentCAD converts one or more hand-drawn orthographic views into executable CAD command sequences that compile into editable 3D models.}
    \label{fig:process}
\end{minipage}
\par\vspace{1em}
}]

\begin{abstract}
Freehand orthographic drawings capture early design intent, yet recovering executable sketch--extrude CAD programs from raster front, right, and top views remains challenging: noisy, dimension-free 2D layouts must be reconciled into an ordered construction history.
We introduce Hand2CAD, a benchmark of 215K ABC-derived synthetic orthographic-drawing--CAD program pairs and 1,000 real three-view sets drawn by CAD-trained participants.
\rev{Recovering CAD programs from orthographic drawings requires both cross-view geometric reasoning and alignment between visual features and program representations.
We propose OrthoLatentCAD to address these two challenges.}
It fuses local geometry with global context on canonical planes, exchanges evidence between paired plane lines indexed by the same shared-axis coordinate, and maps the spatial representation to ordered program latents through queries inherited from a frozen CAD autoencoder.
Across the evaluated one- to three-view configurations, OrthoLatentCAD outperforms protocol-matched baselines in procedural accuracy, compiled geometry, and CAD-kernel validity while benefiting from additional views.
Without real-data fine-tuning, it also outperforms the baselines on held-out real drawings.
We will release the dataset, code, and evaluation scripts.
\end{abstract}

\section{Introduction}
\label{sec:intro}
Freehand front, right, and top views provide a lightweight medium for expressing early mechanical-design intent, whereas downstream CAD workflows require editable histories of sketches and feature operations.
This paper studies how to recover an executable sketch--extrude program from such raster drawings.
Unlike final-shape reconstruction, the task must infer not only the part geometry but also the sketch planes, feature parameters, and ordered operations that construct it, so that the result can be compiled and subsequently edited in a CAD system.
This recovery is difficult because hand drawings omit explicit dimensions and standardized drafting conventions, contain stroke and cross-view alignment noise, and distribute complementary evidence across views.
A successful model must therefore reconcile ambiguous multi-view evidence while producing a construction history that is both geometrically faithful and executable by a CAD kernel.

\rev{Existing CAD generation methods learn construction-history priors from symbolic sequences, latent codes, partial CAD states, or 3D observations~\cite{wu2021deepcad,xu2022skexgen,xu2023hierarchical,li2025mamba}.
Multimodal systems extend CAD generation to text and images~\cite{yavartanoo2024text2cad,zhang2024flexcad,li2025cad,alamgencad,chen2025img2cad,chen2025cadcrafter}, while orthographic reconstruction methods recover wireframes, solids, or parametric code from technical drawings~\cite{hu2023plankassembly,niu2025creftcad,wang2025cad2program,joglekar2026ortho2cad}.
More closely related to our task, Drawing2CAD maps vector drawings to sketch--extrude programs~\cite{qin2025drawing2cad}.
For freehand orthographic rasters, recovering such programs requires combining complementary geometric evidence across views and connecting visual features to program representations.
We address these challenges by explicitly modeling shared-axis correspondences across front, right, and top views and aligning the resulting spatial features with a learned program-latent space.}

Beyond differences in input and output protocols, this task exposes two coupled structural mismatches.
First, orthographic views are not an unordered set of images: the front, right, and top views lie on canonical planes, and each plane pair shares one physical axis.
Lines sampled from a plane pair at the same coordinate along their shared axis provide complementary evidence associated with the same coordinate section in 3D, whereas generic token fusion leaves this relation implicit.
\rev{Second, visual features encode spatial appearance, whereas CAD programs describe geometry through ordered construction operations.
Bridging these representations requires a mapping from drawing evidence to a representation suitable for program decoding.
We learn this target representation with a CAD sequence autoencoder and freeze it during visual training, providing a stable program-latent interface for cross-modal alignment.}
These mismatches also affect executability: locally plausible predictions may violate sketch closure, operation order, or feature dependencies and consequently fail to compile.
\rev{The task thus requires both coordinate-consistent spatial reasoning and effective visual-to-program alignment.}

To support a controlled study of this setting, we introduce Hand2CAD, a benchmark pairing orthographic raster drawings with executable sketch--extrude programs.
Its synthetic component contains 215K ABC-derived pairs and supports training and evaluation across one- to three-view input configurations.
Its real component contains 1,000 three-view sets produced by 20 CAD-trained participants and enables evaluation of cross-domain transfer without real-set fine-tuning.
Further construction details and dataset statistics are provided in the supplementary material.

On this benchmark, we propose OrthoLatentCAD, a coordinate- and slot-consistent framework that maps orthographic evidence to the ordered latent interface of a fixed CAD decoder.
A CAD sequence autoencoder first establishes this decoder-compatible latent space.
For visual mapping, Orthographic Spatial Lifting (OSL) combines local geometry with global context on canonical planes, while Orthographic Mamba Propagation (OMP) uses four-direction intra-plane scans and bidirectional exchange between paired plane lines indexed by the same shared-axis coordinate.
Program-Latent Alignment (PLA) then predicts the target program latent through ordered queries inherited from the autoencoder, after which the frozen decoder produces an executable command sequence.
In this way, spatial correspondence is handled before program decoding, while the visual predictor is anchored to the ordered latent interface learned from CAD sequences.

We evaluate OrthoLatentCAD against three protocol-matched baselines, with view-count comparisons spanning one to three views.
It consistently improves procedural accuracy, compiled geometry, and CAD-kernel validity, with further gains as additional views are provided.
Component ablations support OSL, the two OMP propagation paths, and PLA as complementary parts of the framework.
\rev{OrthoLatentCAD also outperforms the baselines on held-out real drawings without real-data fine-tuning.}

The contributions are threefold:
\begin{itemize}
    \item We introduce Hand2CAD, a benchmark for executable CAD-program recovery from freehand orthographic rasters, comprising 215K synthetic pairs and 1,000 real three-view sets.
    \item We propose OrthoLatentCAD, which couples canonical-plane spatial reasoning and shared-axis evidence propagation with an autoencoder-anchored, decoder-compatible latent interface.
    \item Protocol-matched comparisons and component ablations demonstrate improvements in procedural accuracy, compiled geometry, and CAD-kernel validity across view counts, together with stronger transfer to held-out real drawings.
\end{itemize}
 
\section{Related Work}

\subsection{CAD Program Generation and Program Priors}
DeepCAD represents sketch--extrude histories as command--argument sequences~\cite{wu2021deepcad}, and subsequent methods strengthen CAD program priors through codebooks, hierarchical latents, state-space sequence models, diffusion, and controllable generation~\cite{xu2022skexgen,xu2023hierarchical,li2025mamba,zhang2025diffusion,deng2025mamtiff,wang2024vq}.
Recent work further scales these priors to longer construction histories: GeoFusion-CAD represents CAD programs as hierarchical trees and combines structure-aware diffusion with state-space modeling to capture long-range geometric and topological dependencies~\cite{zhou2026geofusioncad}.
Other systems infer CAD structure from point clouds or boundary representations~\cite{dupont2024transcad,khan2024cad,liu2024point2cad,lambourne2021brepnet,xu2024brepgen,guo2022complexgen}, while multimodal methods condition generation on text, scripts, rendered images, image-derived geometry, or natural photographs~\cite{yavartanoo2024text2cad,zhang2024flexcad,li2025cad,alamgencad,chen2025img2cad,chen2025cadcrafter}.

\rev{Building on these program priors, we combine coordinate-aware processing of freehand orthographic rasters with alignment to a pretrained CAD latent space.
The CAD autoencoder remains frozen during visual training, providing a fixed target representation and decoder.}

\subsection{Orthographic Drawing and Sketch-Based CAD Reconstruction}
Classical orthographic reconstruction recovers wireframes or solids from 2D projections through geometric constraints, topology reasoning, and optimization~\cite{wang1993survey,kuo1998reconstruction,liu2001reconstruction,gong2006reconstruction,furferi20102d}; single-view freehand methods similarly infer abstract 3D interpretations from edge--vertex drawings~\cite{lipson1996optimization}.
Recent learned systems use orthographic cues for restricted shape programs, projection reasoning, or geometry reconstruction~\cite{hu2023plankassembly,niu2025creftcad,zhou2026gaussiancad}.

Methods closer to our task reconstruct parametric code from orthographic inputs.
CAD2Program maps CAD-generated orthographic drawings to text-form parametric programs, while Ortho2CAD combines supervised fine-tuning and geometry-grounded reinforcement learning to generate CadQuery code from dimensioned first-angle drawings~\cite{wang2025cad2program,joglekar2026ortho2cad}.
SOV-CAD instead formulates reconstruction as sequential decision making, using target orthographic projections, partial-model projections, and the active sketch as stepwise visual feedback~\cite{feng2026sovcad}.

\rev{Drawing2CAD~\cite{qin2025drawing2cad} maps vector drawings to sketch--extrude programs using separate command and argument decoders.
Its encoder uses masked mean pooling to aggregate drawing tokens into a single global latent representation that conditions both decoders.
Features from the command decoder further guide argument prediction.}

Sketch-based systems instead parse vectorized strokes, segmented features, interaction traces, or primitive-constraint graphs into CAD-related representations~\cite{Sketch2CAD,li2022free2cad,sun2025sketch2seq,li2026luban}, while sketch datasets capture geometric intent before complete CAD constraints are available~\cite{seff2020sketchgraphs,eitz2012humans,camba2022sketch}.

\rev{OrthoLatentCAD organizes raster evidence on canonical planes, exchanges information along shared axes, and maps the resulting spatial features to the ordered latent interface of a frozen CAD decoder.
This design makes cross-view coordinate relations explicit before predicting the complete sketch--extrude program.}

\subsection{Orthographic Spatial Latents and State-Space Propagation}
Tri-plane representations provide compact 3D-aware feature fields~\cite{chan2022eg3d}, while LRM-style models aggregate one or more images through transformer queries or tri-plane tokens~\cite{hong2023lrm,li2024mlrm}.
CADCrafter and Img2CADSeq move latent prediction toward editable geometry through geometry encoders or hierarchical CAD sequence latents~\cite{chen2025cadcrafter,tan2026img2cadseq}.

For long-range propagation, Mamba and Mamba-2 introduce selective state-space sequence modeling~\cite{gu2024mamba,dao2024mamba2}; visual, multidimensional, and volumetric variants extend directional scans to structured feature fields~\cite{zhu2024visionmamba,liu2024vmamba,li2024mamband,cui2025loma,yoshimura2026sfmamba,lyu2025transamba}.
Within CAD generation, Mamba-CAD applies state-space modeling to command sequences, whereas GeoFusion-CAD combines a hierarchical program tree with state-space diffusion for longer construction histories~\cite{li2025mamba,zhou2026geofusioncad}.

In contrast, OMP applies four-direction state-space propagation within canonical orthographic planes and bidirectional exchange between paired lines indexed by the same shared-axis coordinate.
PLA then maps this coordinate-structured field to the fixed decoder's ordered program-latent interface.

\begin{figure}[!t]
    \centering
    \includegraphics[width=\linewidth]{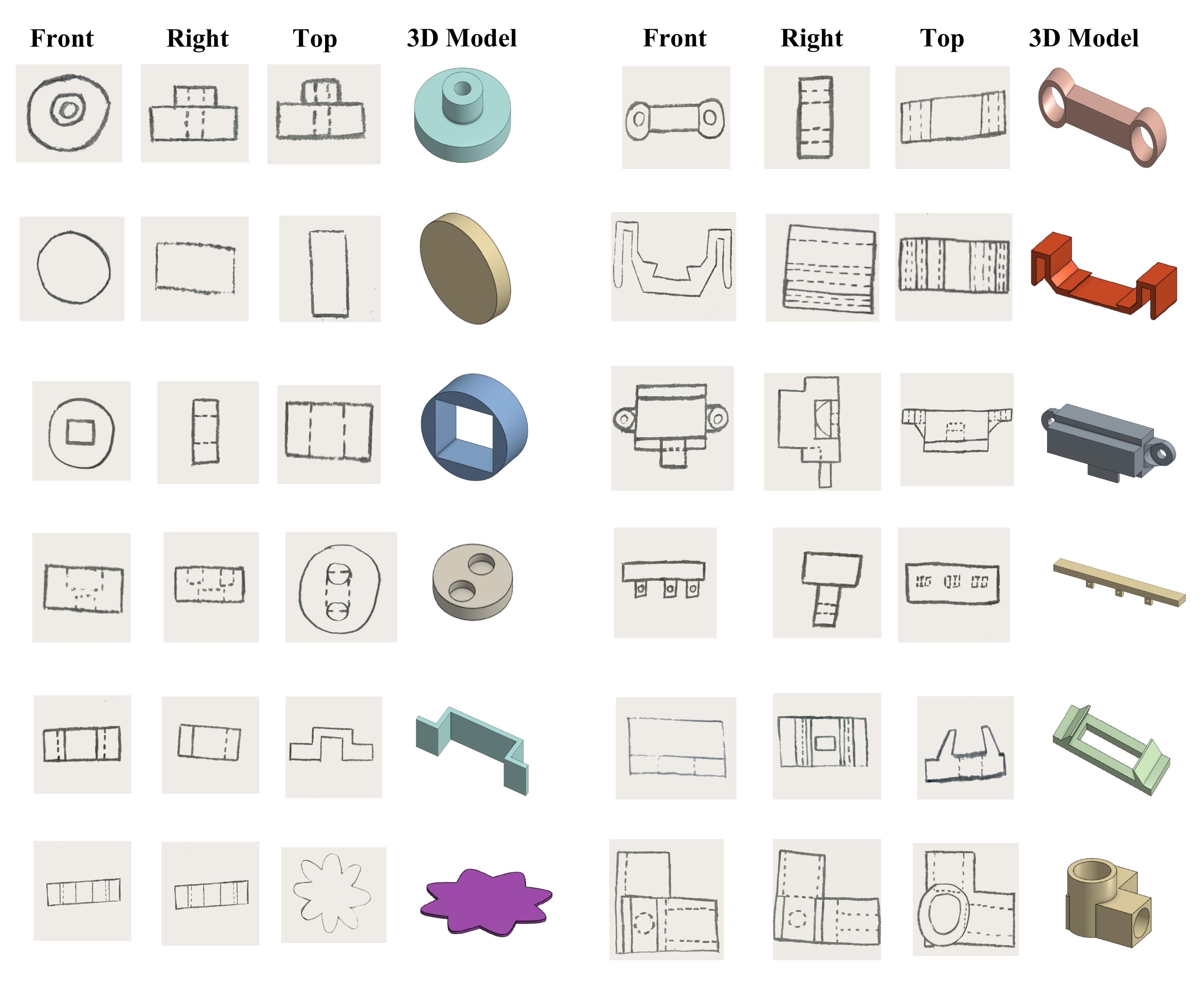}
    \caption{\rev{\textbf{Samples from the Hand2CAD benchmark.} Orthographic drawing views and their corresponding compiled 3D CAD models.}}
    \label{fig:dataset_samples}
\end{figure}

\begin{figure*}[!t]
\centering
\begin{overpic}[width=\textwidth]{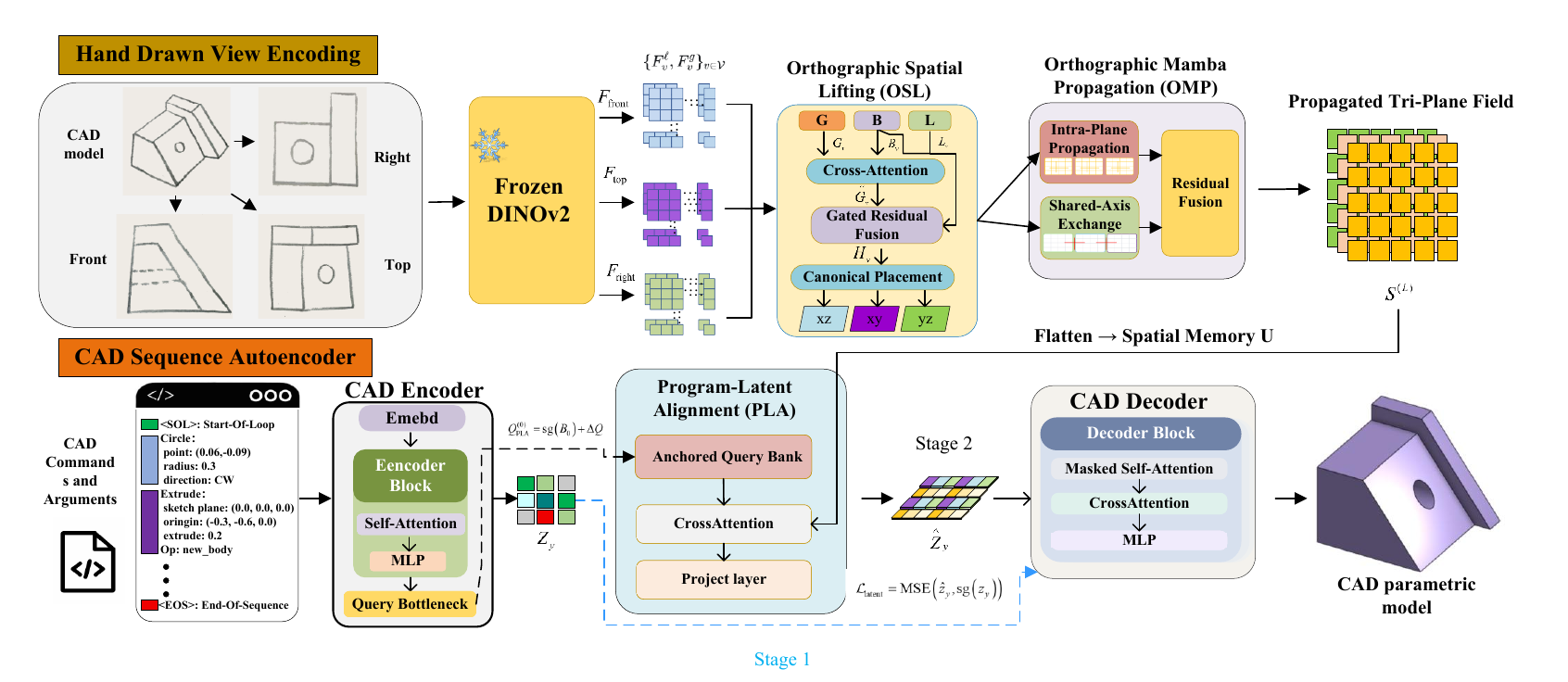}
\put(10.20,9.66){\color{white}\rule{0.080\textwidth}{0.010\textwidth}}
\put(10.42,9.91){\makebox(0,0)[lb]{\fontsize{3.65}{4.1}\selectfont\bfseries\color{black}Origin: (-0.3, -0.6, 0.0)}}
\put(23.55,13.20){\color[HTML]{7B963F}\rule{0.057\textwidth}{0.014\textwidth}}
\put(26.30,13.43){\makebox(0,0)[b]{\fontsize{5.5}{6}\selectfont\bfseries\color{black}Encoder}}
\put(23.30,16.02){\color[HTML]{CEC4DC}\rule{0.058\textwidth}{0.0165\textwidth}}
\put(26.32,16.44){\makebox(0,0)[b]{\fontsize{5.5}{6}\selectfont\bfseries\color{black}Embedding}}
\end{overpic}
\caption{\textbf{Overview of OrthoLatentCAD.} Stage~1 learns an ordered CAD latent space through bottleneck queries $B_0$. In Stage~2, OSL and OMP aggregate orthographic evidence on canonical planes, and PLA anchors visual queries to $B_0$ to predict $\hat z_y$ for the frozen CAD decoder.}
\label{fig:method_main}
\end{figure*}

\section{Problem Setting and Hand2CAD Benchmark}
\label{sec:problem}

\subsection{Task Definition}
Given one or more rasterized orthographic views of a single mechanical part, the task is to recover the corresponding executable sketch--extrude CAD program.
The drawings preserve view-specific contours and approximate cross-view correspondences, but omit the dimensions, sketch planes, construction history, and feature operations needed to specify that program explicitly.
A successful prediction must therefore recover both part geometry and its ordered construction logic rather than only produce a plausible 3D shape.

Let $\mathcal{V}\subseteq\{F,R,T\}$ denote the available front, right, and top views, and let $\mathcal{I}_{\mathcal{V}}=\{I_v\mid v\in\mathcal{V}\}$ denote their raster drawings.
We represent the target program as a command--argument sequence $y=\{(c_t,a_t)\}_{t=1}^{N_y}$, where $N_y$ is the program length, $c_t$ specifies the command type, and $a_t$ contains its quantized argument tokens.
We formulate orthographic-drawing-to-CAD generation autoregressively as:
\begin{equation}
\label{eq:goal}
p_\theta(y\mid\mathcal{I}_{\mathcal{V}})
=\prod_{t=1}^{N_y}
p_\theta(c_t\mid\mathcal{I}_{\mathcal{V}},y_{<t})\,
p_\theta(a_t\mid\mathcal{I}_{\mathcal{V}},y_{<t},c_t).
\end{equation}
Here, $y_{<t}=\{(c_i,a_i)\}_{i=1}^{t-1}$ denotes the preceding command--argument pairs, and $\theta$ denotes the trainable model parameters.

\subsection{Benchmark Overview}
Hand2CAD pairs each orthographic drawing set with its ground-truth executable CAD sequence and kernel-compiled solid, enabling evaluation of procedural accuracy, geometric fidelity, and CAD-kernel validity.
Its synthetic portion contains 215K ABC-derived sketch--CAD pairs represented under the DeepCAD-style sketch--extrude grammar~\cite{koch2019abc,wu2021deepcad}.
For each retained sequence, we compile it into a solid, project the solid into aligned front, right, and top views with FreeCAD~\cite{FreeCAD}, and rasterize the resulting contours as drawing-style inputs.
\rev{For cross-domain evaluation, 20 CAD-trained participants produce 1,000 real three-view drawing sets from distinct CAD programs sampled from the held-out filtered test split. These drawings are used exclusively for evaluation.}
The benchmark fixes the source-model split, command vocabulary, tokenization, and CAD-kernel compilation protocol across all compared methods.
Representative samples are shown in Fig.~\ref{fig:dataset_samples}; detailed filtering criteria, rendering settings, split construction, sequence statistics, real-drawing protocol, and metric definitions are provided in the supplementary material.

\section{OrthoLatentCAD}
\label{sec:method}

\rev{OrthoLatentCAD combines coordinate-aware processing of orthographic drawings with alignment to a pretrained CAD program representation.
As shown in Fig.~\ref{fig:method_main}, Stage~1 learns a CAD sequence autoencoder, whose encoder and decoder are then frozen.
Stage~2 organizes and propagates visual evidence through OSL and OMP, and uses PLA to predict the program latent for the fixed decoder.}
At inference time, the CAD encoder is removed and the predicted latent conditions the fixed CAD decoder.

\subsection{CAD Sequence Autoencoder}

\rev{We train a deterministic CAD sequence autoencoder to define the target program-latent space for cross-modal learning.}
For a command--argument sequence $y=\{(c_t,a_t)\}_{t=1}^{N_y}$, a Transformer encoder produces contextual states $H_y$, and a learnable query bank $B_0\in\mathbb{R}^{K\times C_z}$ reads these states through the bottleneck $B_{\mathrm{cad}}$:
\begin{equation}
\label{eq:cad_autoencoder}
\begin{aligned}
H_y&=E_{\mathrm{seq}}(\operatorname{Emb}(y)),\\
z_y&=B_{\mathrm{cad}}(B_0;H_y)
\in\mathbb{R}^{K\times C_z},\\
\ell_{y,t}&=D_{\mathrm{cad}}(z_y;y_{<t}).
\end{aligned}
\end{equation}
Here, $\ell_y=(\ell_{y,t})_{t=1}^{N_y}$ collects the teacher-forced command and argument logits, while the ordered queries in $B_0$ define $K$ latent slots of width $C_z$.
The autoencoder minimizes the masked command and quantized-argument loss $\mathcal{L}_{\mathrm{seq}}(\ell_y,y)$ defined in the supplementary material:
\begin{equation}
\label{eq:ae_loss}
\mathcal{L}_{\mathrm{AE}}(y)=\mathcal{L}_{\mathrm{seq}}(\ell_y,y).
\end{equation}

After Stage~1, the CAD encoder and decoder are fixed, $z_y$ serves as the supervision target for drawing-conditioned prediction, and a frozen copy of $B_0$ anchors the PLA queries used in Stage~2.
Unlike a single pooled code, the bottleneck preserves $K$ slots while removing dependence on the input sequence length.
\rev{Joint training with the autoregressive decoder establishes the full ordered latent tensor as the interface for program reconstruction.}
\rev{Freezing the CAD modules keeps the target latent representation and decoder fixed during Stage~2.}

\subsection{Orthographic Spatial Lifting}
OSL combines spatially localized intermediate features with deep global context before canonical placement.
For each view $I_v$, intermediate and deep blocks of the frozen DINOv2 encoder provide local-detail and global-context patch features, respectively.
After two-dimensional positional encoding and view-specific projection, let $P_v^{\mathrm{loc}}$ and $P_v^g$ denote their spatial grids.
A spatial residual branch refines $P_v^{\mathrm{loc}}+e_v$ into $L_v$, while pooled Transformer context produces global tokens $G_v$ from $P_v^g+e_v$, where $e_v$ identifies the source view.
\rev{Cross-attention uses the shared spatial reference $B_v$ to query the global tokens. A point-wise $\operatorname{Gate}$ module then predicts two fusion weights at each grid position from the local features and the retrieved global context:}
\begin{equation}
\begin{aligned}
B_v&=\tfrac{1}{2}(P_v^{\mathrm{loc}}+P_v^g)+e_v,\\
\widetilde G_v&=\operatorname{CA}(B_v;G_v),\\
[w_v^{\mathrm{loc}},w_v^g]&=\operatorname{Softmax}\!\left(
\operatorname{Gate}([L_v\Vert\widetilde G_v])\right),\\
H_v&=B_v+w_v^{\mathrm{loc}}\odot s_{\mathrm{loc}}\odot(L_v-B_v)\\
&\quad+w_v^g\odot s_g\odot(\widetilde G_v-B_v).
\end{aligned}
\label{eq:osl_fusion}
\end{equation}
Here, $s_{\mathrm{loc}}$ and $s_g$ are learnable channel-wise scales.
The gated residual update injects global context without overwriting spatially localized contours and endpoints.
We use $\mathcal{P}=\{xy,yz,xz\}$ with $p(F)=xz$, $p(R)=yz$, and $p(T)=xy$.
The deterministic map $\tau_v$ orients $H_v$ on its assigned canonical plane, while a learned token $m_p$ represents an unavailable plane:
\begin{equation}
\label{eq:osl_plane}
S_p^{(0)}=
\begin{cases}
\tau_v(H_v),&p=p(v),\ v\in\mathcal{V},\\
m_p,&p\notin p(\mathcal{V}).
\end{cases}
\end{equation}
Stacking $S_p^{(0)}$ over $p\in\mathcal{P}$ yields $\mathbf S^{(0)}\in\mathbb{R}^{3\times H_s\times W_s\times C_s}$, a common spatial interface for observed and missing views.
\rev{OSL enriches each view before cross-view communication. The deterministic map $\tau_v$ reorients the fused feature grid onto its assigned canonical plane.}

\subsection{Orthographic Mamba Propagation}

\begin{figure}[t]
    \centering
    \includegraphics[width=\linewidth]{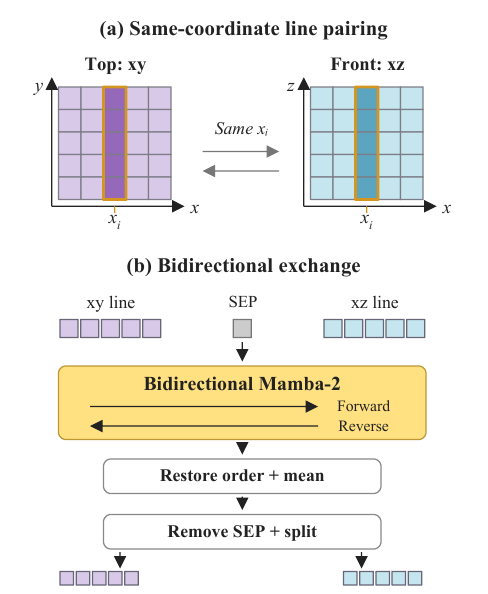}
    \caption{\textbf{Shared-axis exchange in OMP.}
    (a) Fixing $x_i$ selects paired feature lines from the top ($xy$) and front ($xz$) planes, varying along $y$ and $z$.
    (b) The lines, concatenated with a learned separator, undergo bidirectional Mamba-2 propagation before being split and written back.
    The same operation applies to all three shared axes.}
    \label{fig:shared-axis-exchange}
\end{figure}

Each lifted plane contains only the evidence visible in its associated projection.
OMP resolves this incompleteness through four-direction intra-plane propagation and bidirectional exchange between paired plane lines indexed by the same coordinate along their shared axis.
Let $\mathbf{S}^{(\ell)}=\operatorname*{Stack}_{p\in\mathcal{P}}S_p^{(\ell)}$ be the state entering block $\ell$.
For $d\in\mathcal{D}=\{\rightarrow,\leftarrow,\downarrow,\uparrow\}$, $\pi_{p,d}$ scans plane $p$ into a sequence and $\pi_{p,d}^{-1}$ restores its spatial layout:
\begin{equation}
\label{eq:omp_intra}
\bar{S}_{p,d}^{(\ell)}
=\left(
\pi_{p,d}^{-1}\circ
\mathcal{M}_{\mathrm{in}}\circ
\pi_{p,d}
\right)\!\left(S_p^{(\ell)}\right),
\quad (p,d)\in\mathcal{P}\times\mathcal{D},
\end{equation}
where $\mathcal{M}_{\mathrm{in}}$ is the intra-plane Mamba-2 operator.
These directional scans connect distant contour fragments within a view while keeping the three planes separate.
They cannot determine which evidence in another plane refers to the same spatial coordinate, which motivates the constrained cross-plane branch below.

Figure~\ref{fig:shared-axis-exchange} illustrates the shared-axis exchange branch.
Cross-plane propagation follows $\mathcal{A}=\{(xy,xz,x),(xy,yz,y),(xz,yz,z)\}$.
For a plane $p$ containing axis $a$, let $\kappa_p(a)$ return the discrete coordinate index along $a$.
For $(p,q,a)\in\mathcal{A}$, OMP extracts one line from each plane at the same shared-axis coordinate $i$:
\begin{equation}
\label{eq:omp_cross}
\begin{aligned}
L_p^{a,i}&=S_p^{(\ell)}[\kappa_p(a)=i,:],\\
L_q^{a,i}&=S_q^{(\ell)}[\kappa_q(a)=i,:],\\
(\widetilde L_p^{a,i},\widetilde L_q^{a,i})
&=\mathcal{X}_a^{(\ell)}(L_p^{a,i},L_q^{a,i}).
\end{aligned}
\end{equation}
Here, $\mathcal{X}_a^{(\ell)}$ concatenates the paired lines with a learned separator, applies independent forward and reverse Mamba-2 scans, averages their outputs after restoring the reverse order, and separates the two updated lines.
\rev{With a fixed plane concatenation order, bidirectional processing supplies context from both scan directions.}
The $xy/xz$, $xy/yz$, and $xz/yz$ pairs therefore interact at the same $x_i$, $y_i$, and $z_i$, respectively.
\rev{For example, at $x_i$, the $xy$ line varies along $y$ and the $xz$ line varies along $z$, providing complementary plane evidence associated with the coordinate section $x=x_i$.}
The interacted lines are scattered back and averaged over the two axes incident to each plane, yielding $\Delta S_p^{(\ell)}$.
Mean aggregation prevents a plane incident to two shared axes from receiving a systematically larger cross-plane update.
A residual projection with learned LayerScale $\gamma_p^{(\ell)}$ then fuses the original state, four directional features, and cross-plane update:
\begin{equation}
\label{eq:omp_update}
S^{(\ell+1)}_p=S^{(\ell)}_p+
\gamma_p^{(\ell)}\odot\phi_p\!\left([S_p^{(\ell)}\Vert
\operatorname*{Concat}_{d\in\mathcal{D}}\bar S_{p,d}^{(\ell)}
\Vert\Delta S_p^{(\ell)}]\right).
\end{equation}
After $L$ layers, the propagated tri-plane field is
$\mathbf{S}^{(L)}=\mathrm{OMP}(\mathbf{S}^{(0)})$.

\subsection{Program-Latent Alignment, Optimization, and Decoding}

PLA adapts the propagated spatial field to the decoder's ordered latent interface by inheriting the Stage~1 bottleneck queries.
We set $C_s=C_z$, so the frozen query bank can directly initialize the PLA query space without an additional dimensional projection:
\begin{equation}
\label{eq:pla_query}
\begin{aligned}
U&=\operatorname{vec}(\mathbf{S}^{(L)})
\in\mathbb{R}^{3H_sW_s\times C_s},\\
Q^{(0)}=Q_{\mathrm{PLA}}^{(0)}
&=\operatorname{sg}(B_0)+\Delta Q,
\qquad \Delta Q_{\mathrm{init}}=\mathbf{0},\\
Q^{(j+1)}&=\operatorname{CA}_{j}(Q^{(j)};U),
\quad j=0,1,\\
\hat z_y&=\operatorname{Proj}_z(Q^{(2)})\in\mathbb{R}^{K\times C_z}.
\end{aligned}
\end{equation}
Here, $\operatorname{Proj}_z$ is slot-wise, $\operatorname{sg}(B_0)$ is frozen, and $\Delta Q$ provides visual adaptation while retaining the inherited query indices.
Because $B_0$ and $z_y$ use the same $K$-slot order, PLA can compare corresponding latent positions directly.
\rev{The inherited query bank provides the initial slot-wise reference, while $\Delta Q$ adapts the queries to visual inputs.}

In Stage~2, the CAD modules and DINOv2 are frozen, while OSL, OMP, PLA, and $\Delta Q$ are trained through teacher-forced logits $\hat\ell_{y,t}=D_{\mathrm{cad}}(\hat z_y;y_{<t})$:
\begin{equation}
\label{eq:cross_loss}
\begin{aligned}
\mathcal{L}_{\mathrm{latent}}
&=\frac{1}{KC_z}
\left\|\hat z_y-\operatorname{sg}(z_y)\right\|_F^2,\\
\mathcal{L}
&=\lambda_z\mathcal{L}_{\mathrm{latent}}
+\mathcal{L}_{\mathrm{cmd}}
+2\mathcal{L}_{\mathrm{arg}}.
\end{aligned}
\end{equation}
Latent MSE aligns the ordered interface, while command and argument losses require $\hat z_y$ to remain functionally decodable.
The latent-weight schedule $\lambda_z$ emphasizes early alignment and later decoder-visible reconstruction; gradients pass through, but do not update, the fixed decoder.
At inference, teacher forcing is removed and $\hat y=\operatorname{Decode}(D_{\mathrm{cad}},\hat z_y)$ is generated autoregressively; CAD-kernel compilation is used only for evaluation.

\begin{figure*}[t]
\centering
\includegraphics[width=0.91\textwidth]{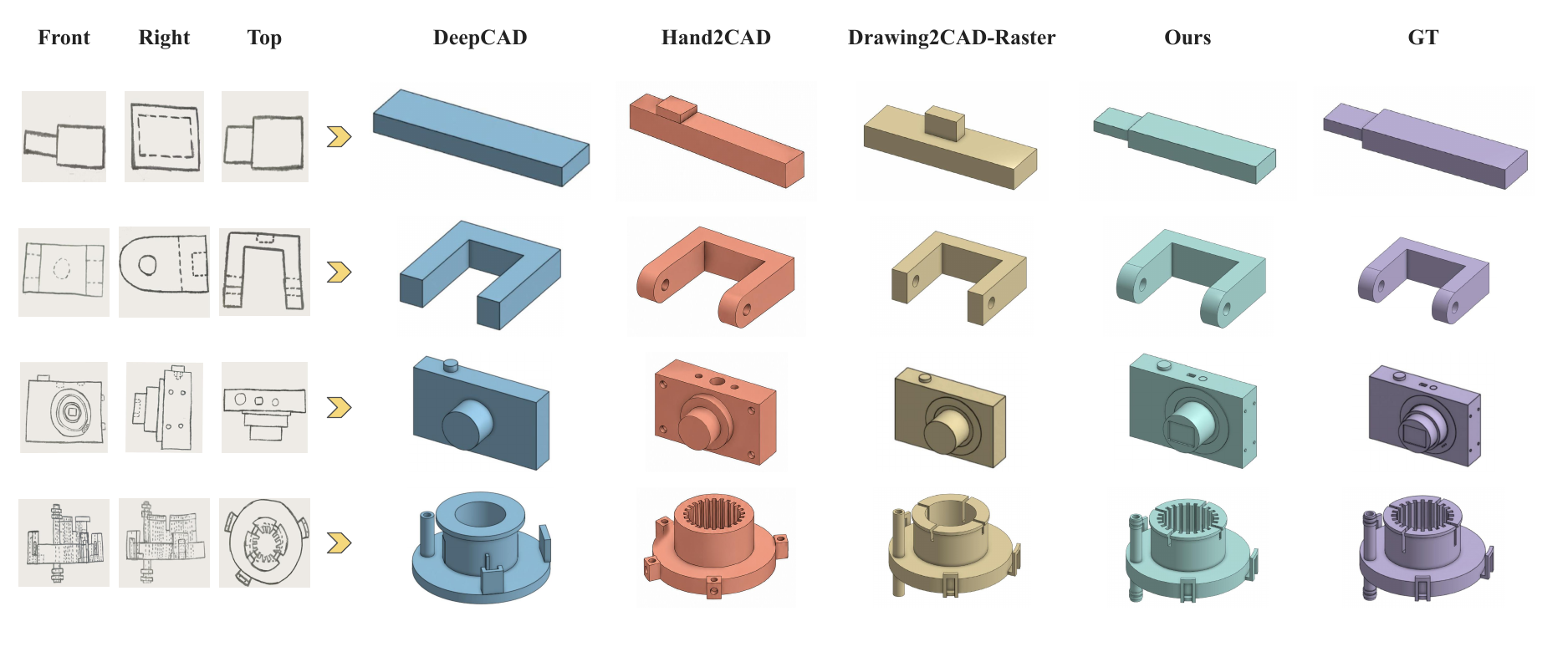}
\caption{\textbf{Qualitative comparison on hand-drawn orthographic inputs.}
Each row shows the front, right, and top views, followed by CAD outputs from DeepCAD, Hand2CAD, Drawing2CAD-Raster, OrthoLatentCAD (Ours), and the ground truth (GT).}
\label{fig:comparison}
\end{figure*}

\begin{table*}[t]
\centering
\resizebox{\textwidth}{!}{
\begin{tabular}{llcccccccccccc}
\toprule
\multirow{2}{*}{Model} & \multirow{2}{*}{Views} & \multicolumn{6}{c}{\rev{Procedural accuracy}} & \multicolumn{3}{c}{Distribution (valid)} & \multicolumn{2}{c}{Instance CD (valid)} & Validity \\
\cmidrule(lr){3-8} \cmidrule(lr){9-11} \cmidrule(lr){12-13} \cmidrule(lr){14-14}
& & ACC$_{cmd}\uparrow$ & ACC$_{param}\uparrow$ & Line$\uparrow$ & Arc$\uparrow$ & Circle$\uparrow$ & Ext.$\uparrow$ & COV$\uparrow$ & MMD$\downarrow$ & JSD$\downarrow$ & Mean$\downarrow$ & Median$\downarrow$ & IR$\downarrow$ \\
\midrule
DeepCAD & F+R+T & 76.4 & 69.2 & 79.4 & 60.3 & 61.2 & 70.1 & 80.1 & 1.72 & 3.98 & 0.82 & 0.53 & 57.2 \\
Hand2CAD & F & 79.8 & 71.6 & 82.4 & 62.1 & 64.2 & 73.3 & 81.7 & 1.66 & 3.80 & 0.78 & 0.49 & 51.3 \\
Hand2CAD & F+R & 81.5 & 73.4 & 83.8 & 64.5 & 67.1 & 75.6 & 82.3 & 1.63 & 3.74 & 0.71 & 0.45 & 48.1\\
Hand2CAD & F+R+T & 83.1 & 75.0 & 85.8 & 67.2 & 69.9 & 78.0 & 82.9 & 1.60 & 3.71 & 0.63 & 0.40 & 42.4 \\
Drawing2CAD-Raster & F+R+T & 84.2 & 74.3 & 87.5 & 69.2 & 72.9 & 80.2 & 83.5 & 1.55 & 3.21 & 0.53 & 0.31 & 36.4 \\
\midrule
OrthoLatentCAD & F & 85.2 & 77.4 & 87.4 & 70.4 & 73.0 & 82.6 & 82.8 & 1.48 & 3.42 & 0.55 & 0.36 & 34.6 \\
OrthoLatentCAD & F+R & 87.3 & 79.2 & 89.3 & 73.2 & 76.1 & 85.0 & 84.6 & 1.39 & 3.28 & 0.49 & 0.32 & 30.8 \\
OrthoLatentCAD & F+R+T & \textbf{89.0} & \textbf{81.0} & \textbf{91.2} & \textbf{76.5} & \textbf{79.3} & \textbf{87.4} & \textbf{86.6} & \textbf{1.31} & \textbf{3.10} & \textbf{0.43} & \textbf{0.28} & \textbf{26.2} \\
\bottomrule
\end{tabular}}
\caption{\rev{\textbf{Main comparison under different view configurations.} F, R, and T denote front, right, and top views. ACC, COV, and IR are percentages; MMD and JSD are scaled by $10^2$, and CD by $10^3$. Geometry metrics use compilable predictions. Drawing2CAD-Raster denotes the raster adaptation described in the supplementary material.}}
\label{tab:main_results}

\end{table*}

\begin{table*}[t]
\centering
\resizebox{\textwidth}{!}{
\begin{tabular}{llcccccccccccc}
\toprule
\multirow{2}{*}{Variant} & \multirow{2}{*}{Views} & \multicolumn{6}{c}{\rev{Procedural accuracy}} & \multicolumn{3}{c}{Distribution (valid)} & \multicolumn{2}{c}{Instance CD (valid)} & Validity \\
\cmidrule(lr){3-8} \cmidrule(lr){9-11} \cmidrule(lr){12-13} \cmidrule(lr){14-14}
& & ACC$_{cmd}\uparrow$ & ACC$_{param}\uparrow$ & Line$\uparrow$ & Arc$\uparrow$ & Circle$\uparrow$ & Ext.$\uparrow$ & COV$\uparrow$ & MMD$\downarrow$ & JSD$\downarrow$ & Mean$\downarrow$ & Median$\downarrow$ & IR$\downarrow$ \\
\midrule
w/o OSL & F+R+T & 85.0 & 77.2 & 87.0 & 70.1 & 72.6 & 82.0 & 81.9 & 1.51 & 3.49 & 0.57 & 0.37 & 35.8 \\
w/o OMP & F+R+T & 86.2 & 78.4 & 88.1 & 71.8 & 74.0 & 83.5 & 83.1 & 1.45 & 3.38 & 0.53 & 0.34 & 32.9 \\
w/o PLA & F+R+T & 84.6 & 76.8 & 86.8 & 69.4 & 72.1 & 81.4 & 81.3 & 1.53 & 3.52 & 0.58 & 0.38 & 37.1 \\
w/o latent loss & F+R+T & 86.5 & 78.5 & 88.4 & 72.3 & 74.6 & 83.9 & 83.8 & 1.44 & 3.35 & 0.52 & 0.33 & 32.6 \\
Intra-plane only & F+R+T & 87.3 & 79.3 & 89.5 & 73.5 & 76.4 & 85.1 & 84.8 & 1.39 & 3.27 & 0.49 & 0.32 & 30.5 \\
Shared-axis only & F+R+T & 86.8 & 78.9 & 89.0 & 72.9 & 75.8 & 84.5 & 84.1 & 1.42 & 3.32 & 0.51 & 0.34 & 31.7 \\
Cross-attn exchange & F+R+T & 88.0 & 80.1 & 90.3 & 74.8 & 77.5 & 86.0 & 85.4 & 1.36 & 3.21 & 0.46 & 0.30 & 28.9 \\
OrthoLatentCAD & F+R+T & \textbf{89.0} & \textbf{81.0} & \textbf{91.2} & \textbf{76.5} & \textbf{79.3} & \textbf{87.4} & \textbf{86.6} & \textbf{1.31} & \textbf{3.10} & \textbf{0.43} & \textbf{0.28} & \textbf{26.2} \\
\bottomrule
\end{tabular}}
\caption{\textbf{Ablation on OrthoLatentCAD components.} All variants use F+R+T. For the \textit{w/o OSL}, \textit{w/o OMP}, and \textit{w/o PLA} controls, the named component is replaced by the protocol-matched alternative described in the text. \rev{Metric definitions and scales follow Table~\ref{tab:main_results}.}}
\label{tab:ablation}
\end{table*}

\section{Experiments and Analysis}
\label{sec:experiments}

Experiments test whether OrthoLatentCAD improves executable CAD recovery across view counts, which components account for the gains, and whether synthetic training transfers to real drawings.
All methods follow the Hand2CAD protocol; extended implementation and metric details are provided in the supplementary material.

\subsection{Evaluation Setup}
\rev{Before rendering, we split the synthetic data by source CAD model into 80\%/10\%/10\% training, validation, and test sets.
The real drawing set is reserved for cross-domain evaluation and is not used for training or fine-tuning.
During OrthoLatentCAD visual training, we use the full synthetic training split and always retain the front view.
The right and top views are independently retained with probability 0.85, with missing views represented by learned mask tokens.
We evaluate the same trained model under F, F+R, and F+R+T inputs, without retraining for individual view configurations.}

We compare against three protocol-matched baselines. Adapted DeepCAD first trains the DeepCAD autoencoder, freezes its latent-to-sequence decoder, and learns a pooled DINOv2-to-latent mapper, whereas Hand2CAD jointly trains a concatenated multi-view conditioner and CAD sequence decoder end to end.
Drawing2CAD-Raster retains Drawing2CAD's encoder and dual-decoder architecture while replacing SVG embeddings with frozen DINOv2 ViT-B/14 patch features.
All methods share the source split, command vocabulary, quantized arguments, and CAD-kernel compilation pipeline.

\rev{\textbf{Implementation details.} In OrthoLatentCAD, input views are resized to $224\times224$. OSL extracts features from blocks 6 and 12 of a frozen DINOv2 ViT-B/14 and projects them to 512 dimensions. We use three $16\times16$ feature grids and four OMP blocks. PLA uses 64 queries and two cross-attention layers. The CAD autoencoder uses a six-layer encoder, a two-layer cross-attention bottleneck, and a six-layer autoregressive decoder. Each training stage is optimized for 200 epochs using AdamW.}

\rev{We evaluate procedural accuracy, CAD-kernel validity, and compiled geometry.
Command and parameter accuracies and IR are evaluated on the full test split, with command-type scores reported separately.
Mean and Median CD measure reconstruction fidelity between valid predictions and their paired targets, whereas COV, MMD, and JSD characterize the generated shape distribution using separately sampled generated and reference sets.
The supplementary material specifies the sampling, normalization, and aggregation procedures.}

\subsection{Quantitative and Qualitative Results}
\rev{\textbf{CAD reconstruction.} Encoding held-out ground-truth programs and decoding autoregressively without teacher forcing yields 99.76\% command accuracy and 98.67\% exact program reconstruction, supporting the fidelity of the program-latent interface. Detailed results appear in the supplementary material.}

Under F+R+T, OrthoLatentCAD exceeds Hand2CAD by 5.9 and 6.0 points in command and parameter accuracy, respectively.
\rev{The largest command-specific gains occur on Arc, Circle, and Extrude, with improvements of 9.3--9.4 percentage points.}
It also raises COV by 3.7 points, reduces Mean CD from 0.63 to 0.43, and lowers IR by 16.2 points.
Together, higher procedural accuracy, better compiled geometry, and lower IR demonstrate improved reconstruction fidelity and executability.

Compared with Drawing2CAD-Raster under F+R+T, OrthoLatentCAD improves command and parameter accuracy by 4.8 and 6.7 points and reduces IR by 10.2 points, \rev{while reducing Mean CD from 0.53 to 0.43.}

Performance also improves monotonically with additional views: from F to F+R to F+R+T, command accuracy increases from 85.2 to 87.3 to 89.0, Mean CD decreases from 0.55 to 0.49 to 0.43, and IR falls from 34.6\% to 30.8\% to 26.2\%.
\rev{OrthoLatentCAD outperforms Hand2CAD at each matched view count.
It also improves procedural accuracy and validity over Drawing2CAD-Raster under F+R+T.}
The supplementary material provides a qualitative view-count comparison that illustrates this progressive recovery and the remaining local errors.

\rev{Figure~\ref{fig:comparison} compares reconstructed parts from the same orthographic inputs.
In the bracket example, OrthoLatentCAD recovers the side openings and rounded ends visible in the ground truth.
The flange example also shows recovery of repeated internal features, although their geometry still differs from the reference.}

\subsection{Ablation Studies}
Table~\ref{tab:ablation} evaluates four component controls and three propagation variants under F+R+T.
In \textit{w/o OSL}, canonical lifting is replaced by a linear projection into three generic grids; in \textit{w/o OMP}, directional and shared-axis propagation is replaced by plane-local residual MLPs; and \rev{in \textit{w/o PLA}, token pooling followed by a shared MLP produces a $64\times512$ latent tensor, retaining the same frozen CAD-encoder target and decoder.}
\textit{w/o latent loss} sets $\lambda_z=0$ and trains the visual pathway only through sequence reconstruction.
The remaining controls retain only intra-plane propagation, retain only shared-axis exchange, or replace shared-axis Mamba-2 with multi-head cross-attention.

\rev{The cross-attention variant operates on the same paired shared-axis line tokens and retains the intra-plane branch.}

OSL and PLA have the largest effects: replacing them lowers command accuracy by 4.0 and 4.4 points and raises IR by 9.6 and 10.9 points, respectively; both replacements also worsen every compiled-geometry metric.
Replacing OMP or removing latent supervision lowers command accuracy by 2.8 or 2.5 points and raises IR by 6.7 or 6.4 points.
\rev{Retaining only intra-plane or shared-axis propagation reduces command accuracy by 1.7 or 2.2 percentage points, respectively, supporting the use of both paths.
Replacing shared-axis Mamba-2 with cross-attention reduces command accuracy by 1.0 percentage point and increases IR by 2.7 percentage points in the evaluated configuration.}

\rev{The PLA and latent-loss ablations support two complementary choices: query-based mapping over the pooling--MLP alternative, and explicit latent supervision alongside command and parameter reconstruction.}

\subsection{Real-Hand Evaluation}
\rev{Table~\ref{tab:real_eval} reports results on 1,000 held-out real three-view drawing sets, without training or fine-tuning on real drawings.}
\rev{Relative to the synthetic F+R+T test set, command and parameter accuracy decrease by 4.8 and 5.0 percentage points, Mean CD increases from 0.43 to 0.62, and IR rises by 13.6 percentage points, showing lower reconstruction accuracy and executability on the real drawing set.}
\rev{On real drawings, OrthoLatentCAD exceeds Drawing2CAD-Raster by 4.4 and 2.9 percentage points in command and parameter accuracy, respectively.
It also reduces Mean CD from 0.72 to 0.62 and IR from 43.5\% to 39.8\%.
These results show that OrthoLatentCAD's reconstruction advantage persists on the held-out real set.} 

\begin{center}
\begin{minipage}{\columnwidth}
\centering
{\footnotesize
\setlength{\tabcolsep}{2pt}
\begin{tabular*}{\linewidth}{@{\extracolsep{\fill}}lcccc@{}}
\toprule
Model & ACC$_{cmd}\uparrow$ & ACC$_{param}\uparrow$ & CD$_{\rm valid}\downarrow$ & IR$\downarrow$ \\
\midrule
DeepCAD & 62.3 & 54.1 & 0.95 & 55.2 \\
Hand2CAD & 78.4 & 69.7 & 0.75 & 48.7 \\
Drawing2CAD-Raster & 79.8 & 73.1 & 0.72 & 43.5 \\
OrthoLatentCAD & \textbf{84.2} & \textbf{76.0} & \textbf{0.62} & \textbf{39.8} \\
\bottomrule
\end{tabular*}}
\captionof{table}{\rev{\textbf{Real hand-drawn evaluation.} F+R+T results on 1,000 held-out three-view drawing sets without real-set fine-tuning. ACC and IR are percentages; CD denotes mean CD over compilable predictions, scaled by $10^3$ as in Table~\ref{tab:main_results}. Additional results are provided in the supplementary material.}}
\label{tab:real_eval}
\end{minipage}
\end{center}

\section{Limitations}
The current benchmark and model are limited to single-part sketch--extrude programs and assume labeled, approximately aligned front, right, and top views.
They do not yet cover broader CAD grammars, assemblies, arbitrary perspective drawings, or unconstrained designer sketches.
\textcolor{black}{The real-hand evaluation measures reconstruction from drawings collected from CAD-trained participants under a controlled protocol.}
Finally, because sketch loops are defined by numerical arguments, small prediction errors may break endpoint closure or introduce invalid intersections, causing profile construction or extrusion to fail in FreeCAD.
Future work will extend the program grammar and drawing conditions, diversify real-user evaluation, and incorporate constraint-aware decoding or kernel-guided repair.

\section{Conclusion}
We introduced the Hand2CAD benchmark for recovering executable CAD programs from freehand orthographic drawings.
We also proposed OrthoLatentCAD to bridge coordinate-structured view evidence with the ordered latent interface of a fixed CAD decoder.
Its canonical-plane lifting, shared-axis propagation, and program-latent alignment improve procedural accuracy, compiled geometry, and CAD-kernel validity over protocol-matched baselines.
\rev{Additional views improve reconstruction, and the gains transfer to held-out real drawings without fine-tuning.}

\FloatBarrier
\clearpage
\section*{\texorpdfstring{\textcolor{black}{LLM Usage Statement}}{LLM Usage Statement}}
\textcolor{black}{OpenAI Codex was used solely for language editing and improving the clarity of the manuscript. The authors reviewed and approved all revisions.}

{\small
\bibliographystyle{ieeenat_fullname}
\bibliography{main}
}

\end{document}


\maketitlesupplementary

\section{Dataset Construction and Real-Hand Protocol}
This section supplements the Hand2CAD benchmark overview in the main paper with details of the synthetic orthographic-drawing--CAD pipeline, the held-out real-hand collection protocol, and dataset statistics.
These details support benchmark reproduction and characterize the program-length distribution and command imbalance relevant to the evaluation.

\subsection{Synthetic Sketch--CAD Pairs}
We construct the synthetic portion of Hand2CAD from FeatureScript-based construction histories of public Onshape documents indexed by ABC~\cite{koch2019abc}.
Following the DeepCAD parsing protocol~\cite{wu2021deepcad}, we retain histories expressible by the sketch--extrude grammar and represent each as an executable command--argument sequence $y=\{(c_t,a_t)\}_{t=1}^{N_y}$.
Each sequence is compiled into a solid with FreeCAD~\cite{FreeCAD} and projected into aligned front, right, and top views.
Because both the solid and the projections originate from the same sequence, every drawing set remains paired with its exact executable target.

The rendering pipeline converts each compiled solid into pencil-style orthographic views while preserving its executable CAD target.
After kernel compilation, visible contour curves are extracted from each canonical view, normalized to a centered square canvas, and rasterized as rounded black strokes.
Each view is rendered at $512\times512$ resolution with a default stroke width of 1.8 pixels before being resized for network input.
When sketch-style augmentation is enabled, we apply smooth local perturbations to the discretized contour points and introduce mild appearance variations to approximate stroke irregularity, pressure variation, and small contour deviations in hand-drawn references.
Because these perturbations are applied after geometric projection, the resulting drawing remains anchored to the original CAD sequence.

To ensure that every retained pair provides a valid learning target, we apply quality control at both the program and view levels.
We discard a sample if its command sequence cannot be compiled by the CAD kernel or falls outside the filtered sketch--extrude vocabulary, or if a projected view is empty, severely cropped, or extends beyond the image frame.
After filtering, the synthetic benchmark contains 215K paired sketch--CAD examples.
We split the CAD source models, rather than their rendered or augmented variants, into 80\%/10\%/10\% training, validation, and test subsets, preventing samples derived from the same source model from crossing subsets.

\begin{figure*}[tbp]
\centering
\includegraphics[width=\linewidth]{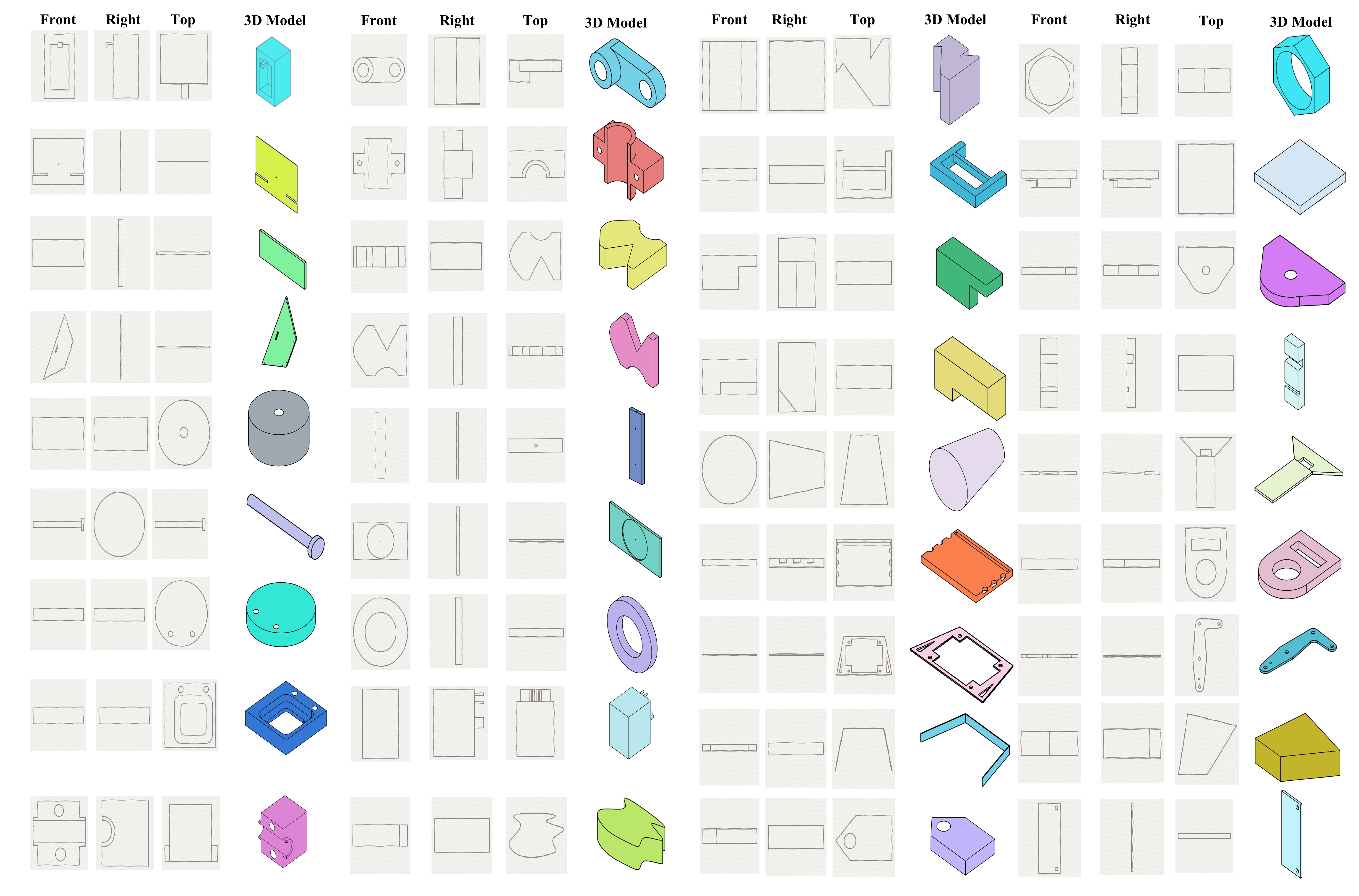}
\caption{Examples from the Hand2CAD benchmark: orthographic raster views and corresponding compiled 3D CAD models.}
\label{fig:supp_3xsvg-v2}
\end{figure*}

\subsection{Real-Hand Evaluation Set}
The real-hand set evaluates cross-domain transfer under a source-disjoint protocol.
We sample 1,000 distinct CAD programs without replacement from the held-out filtered test split and assign them evenly to 20 CAD-trained mechanical-design participants, with 50 drawing tasks per participant.
For each task, the assigned program is compiled into a 3D solid and rendered as a reference from which the participant manually produces the corresponding front, right, and top views.
This procedure yields 1,000 participant--part three-view drawing sets that are used exclusively for evaluation; no real drawings are used for training or fine-tuning.
All participants provide informed consent for the research use, publication, and public release of their anonymized drawings, and the dataset contains no identifying participant information.

\begin{figure*}[tbp]
\centering
\includegraphics[width=\linewidth]{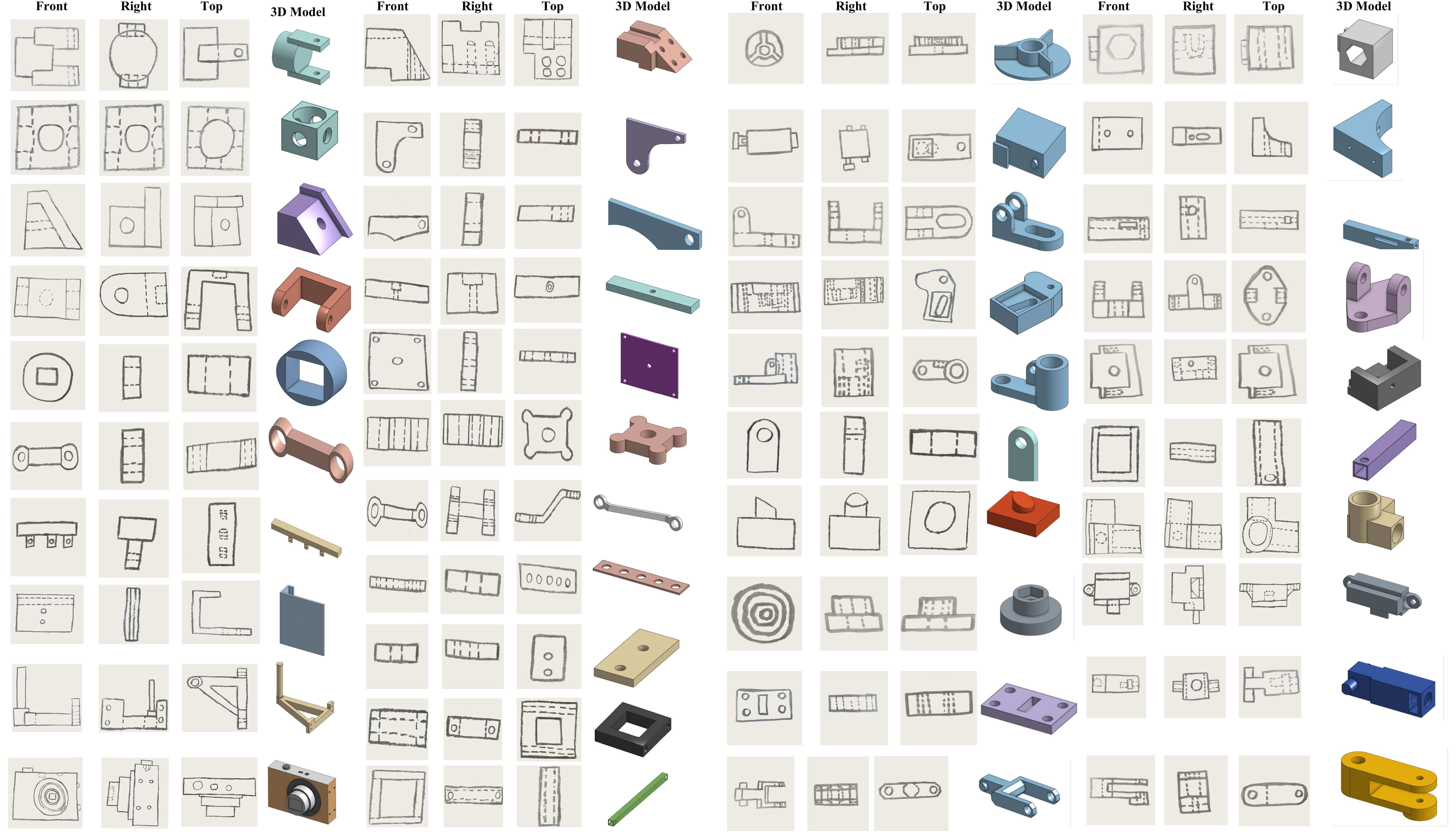}
\caption{Representative real hand-drawn orthographic three-view samples used for cross-domain evaluation.}
\label{fig:supp_hand2sketch}
\end{figure*}

\subsection{Dataset Statistics}
\begin{figure*}[tbp]
  \centering
  \subcaptionbox{Sequence length distribution\label{fig:supp_data_length}}{
    \includegraphics[width=0.48\linewidth]{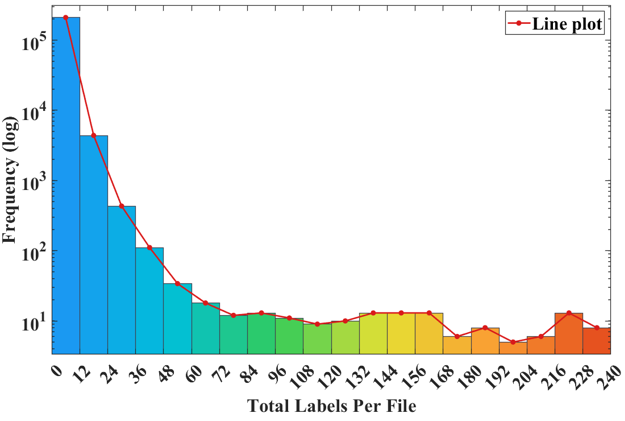}}
  \hfill
  \subcaptionbox{Command-type composition\label{fig:supp_data_command}}{
    \includegraphics[width=0.48\linewidth]{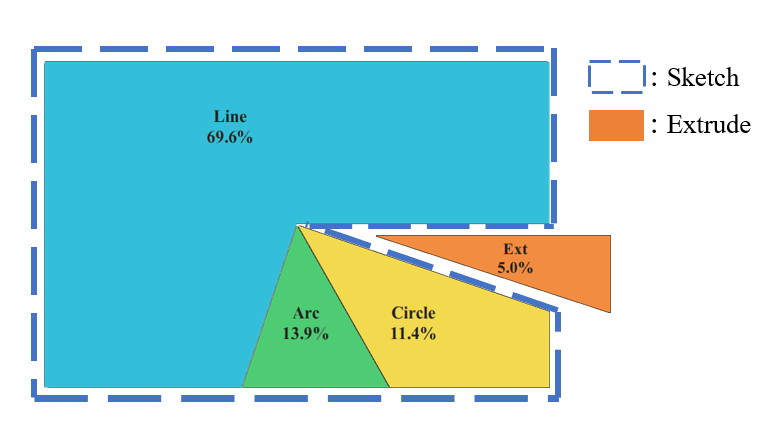}}
  \caption{Dataset statistics of Hand2CAD. (a) Distribution of token-sequence lengths. (b) Command-type composition.}
  \label{fig:supp_dataset_stats}
\end{figure*}

Figure~\ref{fig:supp_dataset_stats} summarizes the program-length distribution and command composition of the filtered synthetic benchmark.
The raw token-sequence lengths are strongly right-skewed: most programs fall within the shortest ranges, while a sparse tail extends to 240 tokens, reflecting a smaller subset with longer construction histories.
Lines account for 69.6\% of commands, followed by arcs at 13.9\%, circles at 11.4\%, and extrusions at 5.0\%.
Together, these distributions show that Hand2CAD combines substantial command imbalance with a broad range of procedural complexity, requiring models to handle both frequent sketch primitives and relatively rare feature-level operations.

\section{CAD Program Representation}
All evaluated systems use the same DeepCAD-style program representation so that performance differences reflect their conditioning architectures rather than different output grammars.
As summarized in Table~\ref{tab:supp_deepcad_token}, continuous CAD parameters are uniformly quantized into integer tokens in $[11,266]$, while values from 0 to 10 are used according to token type for padding, sequence initialization, sketch hierarchy, extrusion termination, Boolean-operation categories, and binary flags~\cite{wu2021deepcad,khan2024cad}.
\rev{This vocabulary and tokenization are shared by all evaluated methods.}

\subsection{Sketch Tokenization}
Each sketch is serialized as a four-level hierarchy comprising the sketch, sketch planes, loops, and curve segments.
A sketch plane contains one or more closed loops, where each loop is represented either by a circle or by an ordered chain of connected line and arc segments.
Curve geometry is encoded using discrete two-dimensional coordinate pairs $(p_x,p_y)$ that specify endpoints or control points on the sketch plane.

To preserve this hierarchy after flattening it into an autoregressive sequence, we insert four structural markers:
\begin{itemize}
    \item $b_s$: start of a sketch,
    \item $b_f$: start of a sketch plane,
    \item $b_l$: start of a loop,
    \item $b_c$: start of a curve.
\end{itemize}

These markers make the boundaries between sketches, planes, loops, and curves explicit without requiring a separate hierarchical decoder.

\subsection{Extrusion Tokenization}

Each extrusion block specifies how a sketched profile is positioned, scaled, swept, and combined with the current solid.
In the global coordinate frame, its continuous parameters encode translation $(\tau_x,\tau_y,\tau_z)$, orientation $(\theta,\phi,\gamma)$, sketch scale $\sigma$, and the forward and backward extrusion distances $d^{+}$ and $d^{-}$.
The categorical token $\beta$ specifies one of four Boolean roles: creating a new body, removing material, performing a union, or forming an intersection.
The termination token $e_e$ marks the end of the extrusion block, and all continuous parameters use the same quantization range as the sketch geometry.

\subsection{Sequence Assembly}

After tokenization, a complete CAD program is organized as an ordered sequence of $n_e$ sketch--extrude construction steps:
\[
\begin{aligned}
\mathcal{C}&=(C_j)_{j=1}^{n_e},
& C_j&=(t_{j,k})_{k=1}^{m_j},\\
t_{j,k}&\in\mathcal{T},
& |\mathcal{T}|&=d_t.
\end{aligned}
\]
Here, $C_j$ contains the sketch hierarchy and extrusion parameters of construction step $j$, $m_j$ is its token length, and $\mathcal{T}$ is the token vocabulary of size $d_t$.
The token $e_e$ terminates an individual extrusion block, whereas the end-of-sequence command terminates the complete CAD program.
A \texttt{cls} token is prepended to initialize decoding, and \texttt{pad} tokens align variable-length programs within a batch.
The end-of-sequence symbol is predicted by the autoregressive command head and therefore does not require a duplicate argument ID in Table~\ref{tab:supp_deepcad_token}.
Flattening the ordered construction steps yields the command--argument sequence $y=\{(c_t,a_t)\}_{t=1}^{N_y}$ used in the main paper.

\begin{table*}[tbp]
\centering
\resizebox{\textwidth}{!}{
\begin{tabular}{c|c|c|c|c}
\hline
\textbf{Sequence Type} & \textbf{Token Type} & \textbf{Token Value} & \textbf{Token Representation} & \textbf{Description} \\ \hline
\multirow{10}{*}{\begin{tabular}[c]{@{}c@{}}Sketch\\ Sequence\end{tabular}} 
& pad & 0 & $(0, 0)$ & Padding Token \\ 
& cls & 1 & $(1, 0)$ & Begin-of-Sequence Token \\ \cline{2-5} 
& $b_s$ & 2 & $(2, 0)$ & Start Sketch \\ 
& $b_f$ & 3 & $(3, 0)$ & Start Sketch Plane \\ 
& $b_l$ & 4 & $(4, 0)$ & Start Loop \\ 
& $b_c$ & 5 & $(5, 0)$ & Start Curve \\ \cline{2-5}
& $(p_x,p_y)$ & $[11...266]^2$ & $(p_x, p_y)$ & Coordinates \\ 
& $\alpha$ & $[11...266]$ & $(\alpha, 0)$ & Arc Parameter (Alpha) \\
& $f$ & $\{0, 1\}$ & $(f, 0)$ & Arc Parameter (Flip) \\
& $r$ & $[11...266]$ & $(r, 0)$ & Circle Parameter (Radius) \\ \hline
\multirow{11}{*}{\begin{tabular}[c]{@{}c@{}}Extrusion\\ Sequence\end{tabular}} 
& $d_+$ & $[11...266]$ & $(d_+, 0)$ & Extrusion Distance (Positive) \\  
& $d_-$ & $[11...266]$ & $(d_-, 0)$ & Extrusion Distance (Negative) \\ \cline{2-5} 
& $\tau_x$ & $[11...266]$ & $(\tau_x, 0)$ & Translation (x) \\ 
& $\tau_y$ & $[11...266]$ & $(\tau_y, 0)$ & Translation (y) \\ 
& $\tau_z$ & $[11...266]$ & $(\tau_z, 0)$ & Translation (z) \\ \cline{2-5}
& $\theta$ & $[11...266]$ & $(\theta, 0)$ & Orientation (Roll) \\  
& $\phi$ & $[11...266]$ & $(\phi, 0)$ & Orientation (Pitch) \\ 
& $\gamma$ & $[11...266]$ & $(\gamma, 0)$ & Orientation (Yaw) \\  \cline{2-5} 
& $\sigma$ & $[11...266]$ & $(\sigma, 0)$ & Sketch Scaling Factor \\  
& $\beta$ & $\{7,8,9,10\}$ & $(\beta, 0)$ & Boolean Operation Type \\ 
& $e_e$ & 6 & $(6, 0)$ & End Extrusion \\ \hline
\end{tabular}}
\caption{DeepCAD sequence representation used in our experiments. Continuous parameters are uniformly quantized into $[11, 266]$.}
\label{tab:supp_deepcad_token}
\end{table*}

\section{Implementation and Baseline Details}

\subsection{CAD Sequence Autoencoder}
The CAD sequence autoencoder establishes the fixed-shape program-latent interface used by OrthoLatentCAD.
Trained only on ground-truth CAD programs, it maps each variable-length command--argument sequence to a deterministic latent and reconstructs the same DeepCAD-style representation.
\textcolor{black}{It provides stable cross-modal supervision and a fixed executable decoder.}

For a program $y$ of length $N_y$, the token embedding and Transformer sequence encoder produce contextual states
\begin{equation}
H_y=E_{\mathrm{seq}}(\operatorname{Emb}(y))
\in\mathbb{R}^{N_y\times C_z}.
\end{equation}
Here, $C_z$ is the program-latent width.
To remove the dependence on sequence length, a learnable query bank $B_0\in\mathbb{R}^{K\times C_z}$ compresses $H_y$ into $K$ ordered latent slots through $M_b$ cross-attention layers:
\begin{equation}
\begin{aligned}
B^{(0)}&=B_0,\\
B^{(m+1)}&=\operatorname{CA}^{\mathrm{cad}}_m
\!\left(B^{(m)};H_y,H_y\right),\\
z_y&=\operatorname{LN}\!\left(B^{(M_b)}\right)
\in\mathbb{R}^{K\times C_z}.
\end{aligned}
\label{eq:supp_cad_bottleneck}
\end{equation}
Here, $B^{(m)}$ supplies the queries, while $H_y$ supplies the keys and values.
The autoregressive CAD decoder reconstructs the command--argument sequence from $z_y$.
Because the same $K$ query indices are used for every program, $z_y$ provides a consistent fixed-slot target for regression from the visual pathway.
After Stage~1, the CAD encoder, decoder, and query bank $B_0$ are frozen; PLA uses a frozen copy of $B_0$ to anchor the ordered visual queries that predict the same latent interface.

\subsection[Stage~1 Reconstruction Evaluation]{\rev{Stage~1 Reconstruction Evaluation}}
After training is complete, we evaluate the frozen Stage~1 CAD autoencoder on the full held-out Stage~1 test split.
Each ground-truth program is used only as input to the CAD encoder, and the decoder then generates the complete command--argument sequence autoregressively from the resulting latent and the start token, without teacher-forcing prefixes.
\rev{This diagnostic evaluates the reconstruction fidelity of the fixed program-latent interface.} Neither a ground-truth CAD program nor its encoder latent is provided to the drawing-conditioned model during Stage~2 inference.

\paragraph{Metrics.}
We follow the main paper for position-level command accuracy (ACC$_{cmd}$), parameter accuracy (ACC$_{param}$), per-command accuracy, and invalidity ratio (IR).
ACC$_{param}$ is computed over valid argument slots at positions where the command is correctly predicted.
Line, Arc, Circle, and Extrude report the percentage of reference positions of each command type that are correctly classified.
For stricter sequence-level evaluation, Exact Cmd. requires the complete command sequence, including EOS, to match the reference, whereas Exact Prog. additionally requires all valid quantized arguments to match.
\rev{IR is the percentage of Stage~1 reconstructions that fail CAD-kernel execution over the full Stage~1 evaluation set.
IR$_{\mathrm{GT\text{-}valid}}$ measures the same reconstruction failure rate on the subset whose reference programs pass the kernel check.}

\begin{table*}[t]
\centering
\resizebox{\textwidth}{!}{
\begin{tabular}{lcccccccccc}
\toprule
\multirow{2}{*}{Setting} & \multicolumn{6}{c}{Procedural accuracy} & \multicolumn{2}{c}{Exact reconstruction} & \multicolumn{2}{c}{Kernel validity} \\
\cmidrule(lr){2-7} \cmidrule(lr){8-9} \cmidrule(lr){10-11}
& ACC$_{cmd}\uparrow$ & ACC$_{param}\uparrow$ & Line$\uparrow$ & Arc$\uparrow$ & Circle$\uparrow$ & Extrude$\uparrow$ & Exact Cmd.$\uparrow$ & Exact Prog.$\uparrow$ & IR$\downarrow$ & IR$_{\mathrm{GT\text{-}valid}}\downarrow$ \\
\midrule
Stage~1 (free-running) & 99.76 & 99.89 & 99.86 & 99.51 & 99.75 & 99.59 & 99.64 & 98.67 & 2.78 & 0.18 \\
\bottomrule
\end{tabular}}
\caption{\textbf{Free-running Stage~1 reconstruction on held-out CAD programs.} The decoder receives only the latent obtained by encoding each ground-truth program and generates autoregressively without teacher forcing. All values are percentages.}
\label{tab:supp_stage1_upper_bound}
\end{table*}

As shown in Table~\ref{tab:supp_stage1_upper_bound}, the decoder retains at least 99.51\% accuracy across all command types and exactly reconstructs 98.67\% of complete programs despite free-running generation.
Although the raw IR is 2.78\%, only 0.18\% of reference-valid programs become invalid after reconstruction.
\rev{These results show that the frozen CAD autoencoder preserves CAD program information with high reconstruction fidelity under free-running decoding.}

\subsection{OrthoLatentCAD Training}
Stage~2 trains the visual pathway to reproduce the decoder-compatible latent defined by the frozen CAD autoencoder.
Each available orthographic view is encoded by a shared, frozen DINOv2 backbone~\cite{oquab2023dinov2}.
Orthographic Spatial Lifting combines intermediate local evidence with deep global context on canonical planes, Orthographic Mamba Propagation exchanges evidence between cross-plane slices aligned at the same coordinate along their shared axis, and Program-Latent Alignment predicts $\hat z_y$ from queries anchored to the Stage~1 bottleneck interface.
The CAD encoder, CAD decoder, visual backbone, and Stage~1 query bank remain fixed during this stage; OSL, OMP, PLA, and the visual query residual are trainable.

Following DeepCAD~\cite{wu2021deepcad}, sequence reconstruction jointly supervises command types and their quantized arguments.
For $N_c$ valid command positions and $N_p$ argument slots per command, we separate the two decoder-visible terms:
\begin{equation}
\begin{aligned}
\mathcal{L}_{\mathrm{cmd}}
&=\frac{1}{N_c}\sum_{i=1}^{N_c}
\operatorname{CE}(\hat{\mathbf c}_i,c_i),\\
\mathcal{L}_{\mathrm{arg}}
&=\frac{1}{M_a}
\sum_{i=1}^{N_c}\sum_{j=1}^{N_p}
m_{i,j}\,\operatorname{CE}(\hat{\mathbf a}_{i,j},a_{i,j}),\\
M_a&=\sum_{i=1}^{N_c}\sum_{j=1}^{N_p}m_{i,j},
\end{aligned}
\label{eq:supp_seq_loss}
\end{equation}
where $\operatorname{CE}(\cdot,\cdot)$ is cross-entropy, $c_i$ and $a_{i,j}$ are target classes, and $\hat{\mathbf c}_i$ and $\hat{\mathbf a}_{i,j}$ are their predicted logit vectors.
The command-dependent mask $m_{i,j}$ equals one only when argument slot $j$ is valid for command $c_i$; padding commands and unused argument slots are excluded.
The reconstruction objective is $\mathcal{L}_{\mathrm{seq}}=\mathcal{L}_{\mathrm{cmd}}+2\mathcal{L}_{\mathrm{arg}}$.

At training epoch $e$, the visual latent is additionally aligned with the frozen CAD-encoder target:
\begin{equation}
\begin{aligned}
\mathcal{L}_{\mathrm{latent}}
&=\frac{1}{KC_z}
\left\|\hat{z}_y-\operatorname{sg}(z_y)\right\|_F^2,\\
\mathcal{L}(e)
&=\lambda_z(e)\mathcal{L}_{\mathrm{latent}}
+\mathcal{L}_{\mathrm{cmd}}
+2\mathcal{L}_{\mathrm{arg}}.
\end{aligned}
\label{eq:supp_cross_loss}
\end{equation}
Here, $\operatorname{sg}$ blocks gradients through the frozen CAD encoder.
We use the fixed two-phase schedule
\begin{equation}
\lambda_z(e)=
\begin{cases}
2.0, & 1\le e\le 5,\\
0.5, & e\ge 6.
\end{cases}
\label{eq:supp_latent_schedule}
\end{equation}
This schedule emphasizes latent alignment early and decoder-visible reconstruction thereafter, while the command and argument terms ensure that the predicted latent remains decodable into the correct executable program.

Both stages use the full synthetic training split and are optimized for 200 epochs with AdamW, weight decay $10^{-4}$, and cosine learning-rate decay.
Stage~1 uses a batch size of 1,024, a learning rate of $3\times10^{-4}$, and a 2,000-step linear warm-up; the checkpoint with the lowest validation reconstruction loss is then frozen for Stage~2.
Stage~2 is trained on two NVIDIA RTX A6000 GPUs with a per-GPU batch size of 96 (global batch size 192), a learning rate of $1.5\times10^{-4}$, and a 667-step linear warm-up.
Unless stated otherwise, all OrthoLatentCAD view protocols and ablations use this Stage~2 optimization schedule.
The predefined $\lambda_z(e)$ phase schedule is shared by all view protocols and ablations.
For checkpoint selection, every validation epoch is scored with the fixed teacher-forced objective $0.5\mathcal{L}_{\mathrm{latent}}+\mathcal{L}_{\mathrm{cmd}}+2\mathcal{L}_{\mathrm{arg}}$; using the final latent weight keeps checkpoint scores comparable across the Epoch~6 schedule change.
All training examples are retained; CAD programs longer than 200 tokens are truncated to this limit and terminated with an end-of-sequence token.
\rev{Distribution metrics are averaged over repeated sampling evaluations as described in Section~\ref{sec:geometry_metrics}.}

During Stage~2, the front view is always retained, while the right and top views are independently retained with probability 0.85; unavailable views are represented by learned mask tokens.
\rev{The same trained OrthoLatentCAD model is evaluated under F, F+R, and F+R+T inputs, without retraining for individual view configurations.}
Each input image is resized to $224\times224$; OSL reads the intermediate block-6 and final block-12 features of a frozen DINOv2 ViT-B/14.
OrthoLatentCAD uses 512-dimensional features, $16\times16$ tri-plane grids, four OMP blocks, and 64 PLA queries.
The CAD autoencoder comprises a 6-layer Transformer sequence encoder, a 2-layer cross-attention bottleneck with 64 CAD-latent queries, and a 6-layer autoregressive Transformer decoder; all Transformer components use eight attention heads.
The Stage~1 bottleneck queries and the visual residual $\Delta Q$ therefore have the same ordered $64\times512$ interface; $\Delta Q$ is initialized to zero.

\rev{OrthoLatentCAD uses greedy autoregressive decoding at inference, without teacher forcing.
Generation stops at the first end-of-sequence token or the 200-token limit.
Drawing2CAD-Raster retains the original parallel command and argument prediction architecture.
All predicted programs use the shared CAD representation and are compiled through the same FreeCAD pipeline for geometric and validity evaluation.}

\subsection{Expanded Method Formulation}
This subsection makes the tensor flow underlying the main-paper equations explicit.
The following equations expand its deterministic indexing and learned transformations without introducing components beyond the architecture defined in the main paper.

The symbols $\pi_{p,d}$, $\kappa_p$, $\operatorname{Split}$, and $\operatorname{Scatter}$ denote deterministic indexing or rearrangement operations.
Specifically, $\pi_{p,d}$ serializes plane $p$ according to scan direction $d$, while $\pi_{p,d}^{-1}$ restores the resulting sequence to its original plane coordinates.
$\kappa_p(a)$ returns the coordinate index along shared axis $a$ for locations on plane $p$, so $S_p[\kappa_p(a)=i,:]$ extracts the line indexed by coordinate $i$.
After bidirectional Cross-plane Mamba-2 propagation, $\operatorname{Split}_{pq}$ removes the separator state and separates the remaining sequence into updates for planes $p$ and $q$.
Finally, $\operatorname{Scatter}_{p,a}$ writes all coordinate-indexed line outputs associated with axis $a$ back to their positions on plane $p$.
In contrast to these deterministic operations, $\operatorname{Proj}_z$ is a learned slot-wise projection from the spatial feature width $C_s$ to the CAD latent width $C_z$.

\paragraph{Orthographic spatial lifting.}
Let $\mathcal{P}=\{xy,yz,xz\}$ denote the canonical planes, with the fixed assignments $p(F)=xz$, $p(R)=yz$, and $p(T)=xy$ for the front, right, and top views.
For an available view $v\in\mathcal{V}$, the frozen DINOv2 encoder exposes an intermediate feature grid $F_v^{\mathrm{loc}}$ and a deep feature grid $F_v^g$.
Let $\Omega_v$ denote the patch-grid domain of view $v$. A learned view embedding $e_v$ and a learned two-dimensional positional-encoding grid $E_v^{\mathrm{pos}}$ are added to both streams.
The local branch applies parallel spatial refinement with different dilation scales to preserve fine geometry, while the global branch pools the deep features and refines the resulting context tokens with a Transformer:
\begin{equation}
\begin{aligned}
L_v&=\mathcal{R}_{\mathrm{loc}}(F_v^{\mathrm{loc}}+E_v^{\mathrm{pos}}+e_v),\\
G_v&=\mathcal{R}_{\mathrm{global}}
\!\left(\operatorname{Pool}(F_v^g+E_v^{\mathrm{pos}}+e_v)\right).
\end{aligned}
\end{equation}
OSL then uses global-to-local cross-attention and a gated residual to inject the global context into every local position:
\begin{equation}
H_v=\operatorname{Fuse}(L_v,G_v)
\in\mathbb{R}^{H_s\times W_s\times C_s}.
\end{equation}
Here, $H_s$, $W_s$, and $C_s$ denote grid height, grid width, and feature width. The fusion operator supplies context for local evidence without discarding its spatial layout.
Let $\Omega_p$ denote the grid domain of canonical plane $p$. The deterministic mapping $\tau_v:\Omega_v\rightarrow\Omega_{p(v)}$ reorients the fused grid onto its assigned canonical plane rather than learning a geometric unprojection.
\rev{Shared-axis exchange uses a common coordinate convention across the canonical planes: the same index $x_i$ in the $xy$ and $xz$ planes denotes the same coordinate along the shared $x$ axis, with analogous correspondences for $y$ and $z$.}
The initial plane feature is
\begin{equation}
S_p^{(0)}=
\begin{cases}
\tau_v(H_v),&p=p(v),\ v\in\mathcal{V},\\
m_p,
& p\notin p(\mathcal{V}),
\end{cases}
\end{equation}
where $p(\mathcal{V})=\{p(v)\mid v\in\mathcal{V}\}$ and the learned token $m_p\in\mathbb{R}^{C_s}$ is broadcast across an unavailable plane.
Stacking the three plane tensors gives
\begin{equation}
\mathbf{S}^{(0)}
=\operatorname*{Stack}_{p\in\mathcal{P}}S_p^{(0)}
\in\mathbb{R}^{3\times H_s\times W_s\times C_s}.
\end{equation}
The resulting tri-plane state combines local geometric detail, global visual context, explicit view identity, and canonical coordinates before any cross-plane propagation.

\paragraph{Intra-plane and shared-axis propagation.}
Let $N=H_sW_s$ and $\mathcal{D}=\{\rightarrow,\leftarrow,\downarrow,\uparrow\}$.
For plane $p$ and scan direction $d$, the permutation $\pi_{p,d}$ converts the plane into a length-$N$ sequence, and its inverse restores the processed sequence to the original plane coordinates:
\begin{equation}
\begin{aligned}
X_{p,d}^{(\ell)}
&=\pi_{p,d}(S_p^{(\ell)})+e_p+e_d
\in\mathbb{R}^{N\times C_s},\\
\bar S_{p,d}^{(\ell)}
&=\pi_{p,d}^{-1}\!\left(
\mathcal{M}_{\mathrm{in}}^{(\ell)}
(X_{p,d}^{(\ell)})
\right).
\end{aligned}
\end{equation}
Here, $e_p$ and $e_d$ are learned plane and scan-direction embeddings.
The four passes differ in scan order and direction identity; after inverse permutation, all outputs share the same plane coordinates.
Cross-plane propagation follows the three known shared-axis relations
\begin{equation}
\mathcal{A}=\{(xy,xz,x),(xy,yz,y),(xz,yz,z)\}.
\end{equation}
For a relation $(p,q,a)\in\mathcal{A}$, let $n_a$ denote the number of discrete coordinates along shared axis $a$.
OMP extracts the two slices anchored at each identical coordinate $i$:
\begin{equation}
L_p^{a,i}=S_p^{(\ell)}[\kappa_p(a)=i,:],
\qquad
L_q^{a,i}=S_q^{(\ell)}[\kappa_q(a)=i,:].
\end{equation}
Thus, $xy/xz$ pairs share $x_i$, $xy/yz$ pairs share $y_i$, and $xz/yz$ pairs share $z_i$.
Learned plane, axis, and propagation-direction embeddings identify the role of each token in the paired sequence.
The paired sequence is
\begin{equation}
\begin{aligned}
\bar L_p^{a,i}&=L_p^{a,i}+e_p+e_a+e_{p\rightarrow q},\\
\bar L_q^{a,i}&=L_q^{a,i}+e_q+e_a+e_{q\rightarrow p},\\
s_{pq}^{a,i}&=
[\bar L_p^{a,i}\Vert e_{pq}^{\mathrm{sep}}\Vert\bar L_q^{a,i}].
\end{aligned}
\end{equation}
Bidirectional Cross-plane Mamba-2 propagates information through this paired sequence.
After removing the separator state, the output is divided into updates for the two planes:
\begin{equation}
[\widetilde L_p^{a,i}\Vert\widetilde L_q^{a,i}]
=\operatorname{Split}_{pq}\!\left(
\mathcal{B}_{\mathrm{cross}}^{(\ell)}(s_{pq}^{a,i})
\right).
\end{equation}
For each plane $p$, let $\mathcal{A}_p=\{a\mid\exists q,\,(p,q,a)\in\mathcal{A}\ \text{or}\ (q,p,a)\in\mathcal{A}\}$ denote its incident axes. The coordinate-indexed outputs are written back to a full-plane tensor and averaged across these axes:
\begin{equation}
\begin{aligned}
\Delta S_{p,a}^{(\ell)}
&=\operatorname{Scatter}_{p,a}
\!\left(\{\widetilde L_p^{a,i}\}_{i=1}^{n_a}\right),\\
\Delta S_p^{(\ell)}
&=\frac{1}{|\mathcal{A}_p|}
\sum_{a\in\mathcal{A}_p}\Delta S_{p,a}^{(\ell)},
\qquad |\mathcal{A}_p|=2.
\end{aligned}
\end{equation}
Mean aggregation keeps the magnitude of the cross-plane update independent of the number of incident axes and introduces no additional learned fusion module.
Finally, the original plane state, the four restored intra-plane features, and the averaged cross-plane update are fused by a residual projection $\phi_p:\mathbb{R}^{6C_s}\rightarrow\mathbb{R}^{C_s}$ with a learned LayerScale vector $\gamma_p^{(\ell)}$:
\begin{equation}
S_p^{(\ell+1)}=S_p^{(\ell)}+
\gamma_p^{(\ell)}\odot\phi_p\!\left(
[S_p^{(\ell)}\Vert
\operatorname*{Concat}_{d\in\mathcal{D}}\bar S_{p,d}^{(\ell)}
\Vert\Delta S_p^{(\ell)}]
\right).
\end{equation}
After $L$ blocks, the propagated spatial field is $\mathbf{S}^{(L)}=\operatorname{OMP}(\mathbf{S}^{(0)})$.

\paragraph{Program-latent alignment.}
After $L$ OMP blocks, the three propagated planes are flattened into a spatial memory
\begin{equation}
U=\operatorname{vec}(\mathbf{S}^{(L)})
\in\mathbb{R}^{M\times C_s},
\qquad M=3H_sW_s.
\end{equation}
PLA initializes its ordered visual queries from the frozen Stage~1 bottleneck interface:
\begin{equation}
Q^{(0)}
=\operatorname{sg}(B_0)+\Delta Q,
\qquad
\Delta Q_{\mathrm{init}}=\mathbf{0}.
\end{equation}
Because $C_s=C_z$, the frozen Stage~1 query bank and the visual residual share the same width.
Here, $\operatorname{sg}(B_0)$ preserves the Stage~1 query order, while the zero-initialized residual $\Delta Q\in\mathbb{R}^{K\times C_s}$ adapts that interface to visual evidence.
For one attention head in layer $j$, the query-to-spatial attention is
\begin{equation}
\begin{aligned}
A_j
&=\operatorname{softmax}\!\left(
\frac{(Q^{(j)}W_Q^j)(UW_K^j)^\top}
{\sqrt{d_h}}
\right)(UW_V^j).
\end{aligned}
\end{equation}
Here, $W_Q^j$, $W_K^j$, and $W_V^j$ are the learned query, key, and value projections for the illustrated head, $d_h$ is its query/key width, and softmax is evaluated over the $M$ memory positions.
The outputs from all heads are concatenated and processed by the standard output projection, residual connection, and feed-forward update, which we denote by
\begin{equation}
Q^{(j+1)}
=\operatorname{CA}_j(Q^{(j)};U),
\qquad j\in\{0,1\}.
\end{equation}
After two cross-attention layers, the learned projection $\operatorname{Proj}_z$ converts the query states to the CAD latent width:
\begin{equation}
\hat z_y
=\operatorname{Proj}_z(Q^{(2)})
\in\mathbb{R}^{K\times C_z}.
\end{equation}
The residual $\Delta Q$ is shared across samples and learned jointly with OSL, OMP, and the remaining PLA parameters; $B_0$ remains frozen.
Because $\hat z_y$ has the same ordered shape as the frozen target $z_y$, the two representations can be aligned directly with the slot-wise latent loss.
This anchoring reduces the burden of learning visual-to-latent slot correspondence, but it does not impose or guarantee a fixed semantic label for every slot.

\paragraph{Two-stage optimization and inference.}
Stage~1 jointly trains the CAD encoder, including $B_0$, and the decoder using the sequence-reconstruction objective:
\begin{equation}
\begin{aligned}
(\theta_E^*,\theta_D^*)
&=\arg\min_{\theta_E,\theta_D}
\mathbb{E}_{y}\!\left[
\mathcal{L}_{\mathrm{seq}}\big(\ell_y(\theta_E,\theta_D),y\big)
\right],\\
\ell_y(\theta_E,\theta_D)
&:=\left(D_{\theta_D}(E_{\theta_E}(y);y_{<t})\right)_{t=1}^{N_y}.
\end{aligned}
\end{equation}
Here, $y_{<t}$ denotes the ground-truth prefix used for teacher-forced autoregressive decoding.
After Stage~1, $E_{\theta_E^*}$, $D_{\theta_D^*}$, and the learned bottleneck queries $B_0$ are frozen.

For compactness, let the complete visual pathway be
\begin{equation}
G_{\Theta_{\mathrm{vis}}}(\mathcal{I}_{\mathcal{V}})
=
\operatorname{PLA}\!\left(
\operatorname{OMP}\!\left(
\operatorname{OSL}\!\left(
\{E_{\mathrm{img}}(I_v)\}_{v\in\mathcal{V}}
\right)\right)\right),
\end{equation}
where $\Theta_{\mathrm{vis}}$ contains the local/global lifting branches, view embeddings and mask tokens, OMP blocks, residual queries $\Delta Q$, cross-attention layers, and projection $\operatorname{Proj}_z$.
The DINOv2 backbone $E_{\mathrm{img}}$ remains frozen.
Stage~2 predicts $\hat z_y=G_{\Theta_{\mathrm{vis}}}(\mathcal{I}_{\mathcal{V}})$ and uses the fixed CAD-encoder target $z_y=E_{\theta_E^*}(y)$.
Writing the teacher-forced logits as $\hat\ell_y=\left(D_{\theta_D^*}(\hat z_y;y_{<t})\right)_{t=1}^{N_y}$, the training objective at epoch $e$ is
\begin{equation}
\begin{aligned}
\Theta_{\mathrm{vis}}^*
=\arg\min_{\Theta_{\mathrm{vis}}}
\mathbb{E}_{(\mathcal{I}_{\mathcal{V}},y)}
\Bigg[
&\frac{\lambda_z(e)}{KC_z}
\left\|
\hat z_y-\operatorname{sg}(z_y)
\right\|_F^2\\
&+\mathcal{L}_{\mathrm{cmd}}(\hat\ell_y,y)
+2\mathcal{L}_{\mathrm{arg}}(\hat\ell_y,y)
\Bigg].
\end{aligned}
\end{equation}
At inference, the CAD encoder and teacher-forcing prefixes are removed, so final program evaluation is free-running rather than teacher-forced.
The predicted latent is decoded autoregressively as
\begin{equation}
\hat y
=\operatorname{ARDecode}_{D_{\theta_D^*}}\!\left(
G_{\Theta_{\mathrm{vis}}^*}
(\mathcal{I}_{\mathcal{V}})
\right).
\end{equation}
Thus, inference follows the direct path $\mathcal{I}_{\mathcal{V}}\rightarrow\mathrm{OSL}\rightarrow\mathrm{OMP}\rightarrow\mathrm{PLA}\rightarrow\hat z_y\rightarrow\hat y$.

\subsection{Protocol-Matched Baselines}
We compare OrthoLatentCAD with three visual baselines under the same orthographic-input and CAD-output protocol.
The adapted DeepCAD baseline replaces its original unconditional latent generator with a view-conditioned mapper: DINOv2 descriptors from the available orthographic views are pooled and mapped to the latent interface of a DeepCAD-style decoder.
The Hand2CAD baseline instead concatenates the multi-view visual tokens and directly conditions the CAD sequence decoder, without canonical plane lifting, orthographic latent propagation, or supervision from the frozen CAD-autoencoder latent.
The adapted DeepCAD decoder is first obtained by training the DeepCAD autoencoder on the same synthetic program split; the decoder is then frozen while the view-conditioned mapper is optimized.
The Hand2CAD visual conditioner and CAD sequence decoder are instead trained jointly end to end from drawing inputs to target sequences.

All methods use the same source-model split, command vocabulary, quantized argument representation, maximum decoding length, and FreeCAD compilation protocol; whenever methods are compared at a given view count, they receive the same available views.
The comparison therefore changes how orthographic evidence is organized and delivered to the decoder while keeping the output grammar and evaluation procedure fixed.
Methods that retain different input modalities, output languages, or additional supervision are discussed as related systems; the raster adaptation below brings Drawing2CAD into the orthographic-input and CAD-output setting.

\begin{revision}
\subsection{Drawing2CAD Raster Adaptation}
Drawing2CAD-Raster adapts Drawing2CAD~\cite{qin2025drawing2cad} to front, right, and top raster drawings.
A frozen DINOv2 ViT-B/14 extracts patch features from each $224\times224$ view.
A trainable projection maps these features to the Drawing2CAD encoder width, and view and spatial position encodings are added before the view sequences are concatenated.
The resulting tokens replace the SVG primitive embeddings at the input of the original Transformer encoder.

We retain Drawing2CAD's latent bottleneck, separate command and argument decoders, command-guided parameter generation, and soft-target parameter loss.
The input adapter and CAD prediction modules are trained on the common synthetic split while DINOv2 remains frozen.
Evaluation uses the same available views, CAD command vocabulary, quantized arguments, maximum output length, and compilation pipeline as the other methods.

\rev{Following the original Drawing2CAD training protocol, we train with Adam for 200 epochs using a batch size of 256 and a learning rate of 0.001.
The learning rate increases linearly during the first 2,000 steps.
Transformer dropout is set to 0.1, and the gradient clipping threshold is 1.0.}
\end{revision}

\subsection{Ablation Variant Definitions}
All ablations retain the full model's data split, decoder, output grammar, training schedule, and evaluation protocol, changing only the component stated below.
In \textit{w/o OSL}, the fixed front/right/top-to-$xz/yz/xy$ placement is replaced by a linear projection of the concatenated visual tokens into three generic feature grids with the same shape as the canonical planes; OMP and PLA are unchanged.
In \textit{w/o OMP}, OSL is retained, but the directional and shared-axis Mamba-2 updates are replaced by local residual MLP blocks applied independently to the lifted plane tokens.
\rev{In \textit{w/o PLA}, OSL and OMP are retained, while PLA is replaced by a pooling-based MLP adapter.
The propagated spatial memory $U\in\mathbb{R}^{768\times512}$ is pooled to 64 tokens.
A shared MLP is then applied independently to each token to predict $\hat z_y\in\mathbb{R}^{64\times512}$.
This adapter uses neither inherited program queries nor query-to-spatial cross-attention.
The predicted latent is supervised by the same frozen CAD-encoder target and decoded by the same frozen CAD decoder.
The latent-loss weight schedule, sequence reconstruction objective, and remaining training settings are unchanged.}

The remaining controls isolate the propagation operator and latent supervision.
\textit{Intra-plane only} disables shared-axis exchange while retaining the four directional scans within each plane; \textit{shared-axis only} removes the intra-plane scans and retains bidirectional Mamba-2 exchange along the paired $x$, $y$, and $z$ axes.
\textit{Cross-attention exchange} uses the same paired shared-axis line tokens as the full model but replaces their bidirectional Mamba-2 update with multi-head cross-attention, while leaving the intra-plane branch unchanged.
Finally, \textit{w/o latent loss} sets $\lambda_z=0$, so the visual pathway is trained only through sequence reconstruction by the frozen decoder.

\subsection{DeepCAD Latent Conditioning}
The adapted DeepCAD baseline preserves the DeepCAD-style latent-to-sequence decoder but replaces its unconditional latent prior with a view-conditioned mapper.
For each available view $v\in\mathcal{V}$, a pooled DINOv2 descriptor is computed as
\begin{equation}
g_v=\operatorname{Pool}\!\left(E_{\mathrm{img}}(I_v)\right)
\in\mathbb{R}^{C_s},
\qquad v\in\mathcal{V}.
\end{equation}
The descriptors are averaged over the available views:
\begin{equation}
g_{\mathcal{V}}
=\frac{1}{|\mathcal{V}|}\sum_{v\in\mathcal{V}}g_v.
\end{equation}
A lightweight MLP then maps the result to the decoder's latent interface:
\begin{equation}
z_{\mathrm{base}}
=G_{\mathrm{cond}}(g_{\mathcal{V}})
\in\mathbb{R}^{K\times C_z}.
\end{equation}
The same mapper is used for all view counts, and the fixed DeepCAD decoder generates the CAD command--argument sequence from $z_{\mathrm{base}}$.
This construction provides a controlled visual adaptation of DeepCAD but does not retain view-specific layouts or explicitly model canonical planes, shared axes, or program-query alignment.

\section{Evaluation Protocol Details}
The main paper summarizes the common evaluation setting; here we specify the counting rules, tolerance, and treatment of invalid programs used for every method.

\subsection{Procedural Accuracy and Validity}
Let $N$ be the number of aligned, non-padding command positions across the full test split.
Command accuracy is
\begin{equation}
\operatorname{ACC}_{\mathrm{cmd}}
=\frac{1}{N}\sum_{i=1}^{N}
\mathbb{1}[\hat c_i=c_i].
\end{equation}
The Line, Arc, Circle, and Ext. columns apply the same exact command-match test only to positions whose target belongs to the corresponding command type.
Sequences are aligned position-wise through the target end-of-sequence command.
If a prediction terminates early, the remaining non-padding target positions are treated as command mismatches; if it runs longer, it is truncated to the target length.
Premature or missing end-of-sequence decisions therefore reduce $\operatorname{ACC}_{\mathrm{cmd}}$ without changing its target-defined denominator.

Parameter accuracy follows the DeepCAD rule and is evaluated only when the command type is predicted correctly.
Let $m_{i,j}\in\{0,1\}$ indicate whether argument slot $j$ is active for target command $c_i$, $\chi_i=\mathbb{1}[\hat c_i=c_i]$, and $r_{i,j}$ indicate agreement under the argument-type rule: quantized geometric values use the fixed DeepCAD tolerance $\delta$, whereas categorical values require exact equality.
The reported score is
\begin{equation}
\operatorname{ACC}_{\mathrm{param}}
=\frac{\sum_{i,j}m_{i,j}\chi_i r_{i,j}}
{\sum_{i,j}m_{i,j}\chi_i}.
\end{equation}
The tolerance and argument-type rules are fixed for all methods and are not tuned per model.
Because its denominator contains active argument slots only at command-matched positions, $\operatorname{ACC}_{\mathrm{param}}$ measures parameter quality conditional on correct command prediction rather than end-to-end sequence accuracy; it is therefore interpreted together with $\operatorname{ACC}_{\mathrm{cmd}}$.

Let $S$ be the total number of programs in the full test split and $k_s=1$ if FreeCAD executes program $s$ into a valid solid and $k_s=0$ otherwise.
The Invalid Ratio is $100S^{-1}\sum_{s=1}^{S}(1-k_s)$.
Kernel failure can arise from inconsistent command order, invalid parameters, Boolean-operation failure, or a numerically invalid sketch loop even when individual tokens appear plausible.
Invalid predictions contribute to IR but not to point-cloud metrics because they provide no solid from which a surface can be sampled.

\begin{revision}
\subsection{Geometry Metrics}
\label{sec:geometry_metrics}
We follow the DeepCAD evaluation framework~\cite{wu2021deepcad} to measure paired reconstruction fidelity and agreement between shape distributions.
For point clouds $A$ and $B$, the bidirectional squared Chamfer distance is
\begin{equation}
\begin{aligned}
d_{\mathrm{CD}}(A,B)
&=\frac{1}{|A|}\sum_{a\in A}\min_{b\in B}\|a-b\|_2^2\\
&\quad+\frac{1}{|B|}\sum_{b\in B}\min_{a\in A}\|b-a\|_2^2.
\end{aligned}
\end{equation}

\paragraph{Paired reconstruction.}
For method $m$, let $\mathcal{J}_m$ contain the test indices with compilable predictions.
For each $i\in\mathcal{J}_m$, $P_i$ and $Q_i$ denote surface point clouds from the predicted solid and its corresponding target, processed under the reconstruction-evaluation protocol.
Each paired point cloud contains 3,000 sampled surface points.
Mean CD and Median CD are the mean and median of $d_{\mathrm{CD}}(P_i,Q_i)$ over $\mathcal{J}_m$.
They measure geometric fidelity conditioned on a compilable output and are interpreted together with IR over the full test split.

\paragraph{Shape distributions.}
COV, MMD, and JSD use generated and reference sets sampled separately for each evaluation round.
For method $m$ and round $t$, let $\mathcal{G}_m^{(t)}=\{U_j\}_{j=1}^{N_{\mathrm{gen}}}$ be sampled from the pool of compilable predictions and $\mathcal{R}_m^{(t)}=\{V_k\}_{k=1}^{N_{\mathrm{ref}}}$ from the test reference pool.
Each point cloud is normalized by its maximum absolute coordinate before distribution evaluation.
MMD measures the distance from each reference shape to its nearest generated shape:
\begin{equation}
\operatorname{MMD}_m^{(t)}
=\frac{1}{N_{\mathrm{ref}}}\sum_{k=1}^{N_{\mathrm{ref}}}
\min_{1\le j\le N_{\mathrm{gen}}}d_{\mathrm{CD}}(U_j,V_k).
\end{equation}
Coverage measures the fraction of reference shapes selected as the nearest neighbor of at least one generated shape:
\begin{equation}
\begin{aligned}
\mathcal{N}_m^{(t)}
&=\left\{\arg\min_{1\le k\le N_{\mathrm{ref}}}
 d_{\mathrm{CD}}(U_j,V_k)\right\}_{j=1}^{N_{\mathrm{gen}}},\\
\operatorname{COV}_m^{(t)}&=|\mathcal{N}_m^{(t)}|/N_{\mathrm{ref}}.
\end{aligned}
\end{equation}
For voxel-occupancy distributions $p_{\mathcal{G}}$ and $p_{\mathcal{R}}$, JSD is
\begin{equation}
\begin{aligned}
\operatorname{JSD}_m^{(t)}
&=\tfrac{1}{2}D_{\mathrm{KL}}(p_{\mathcal{G}}\|u)
 +\tfrac{1}{2}D_{\mathrm{KL}}(p_{\mathcal{R}}\|u),\\
u&=\tfrac{1}{2}(p_{\mathcal{G}}+p_{\mathcal{R}}),
\end{aligned}
\end{equation}
where the occupancy distributions are formed on the same voxel grid and logarithms use base two.
The reported distribution metrics are arithmetic means over $T$ evaluation rounds using the fixed prediction pool.

The sampling and metric settings are shared across OrthoLatentCAD, DeepCAD, Hand2CAD, Drawing2CAD-Raster, and all ablations.
Tables report COV as a percentage, MMD and JSD multiplied by $10^2$, and Mean and Median CD multiplied by $10^3$.
\end{revision}

\subsection{Full Real-Hand Metrics}
Table~\ref{tab:supp_real_full} reports the full metric set for the 1,000 real participant--part orthographic three-view drawing sets summarized in the main paper.
The evaluation uses F+R+T inputs without training or fine-tuning on the real drawing set; ACC and IR include all 1,000 examples, while geometry metrics follow the compilable-output definition above.
\rev{On real drawings, OrthoLatentCAD improves COV by 3.5 percentage points over Drawing2CAD-Raster and reduces MMD from 1.62 to 1.48 and Median CD from 0.45 to 0.40.}

\begin{table*}[tbp]
\centering
\resizebox{\textwidth}{!}{
\begin{tabular}{lcccccccccccc}
\toprule
\multirow{2}{*}{Model} & \multicolumn{6}{c}{\rev{Procedural accuracy}} & \multicolumn{3}{c}{Distribution (valid)} & \multicolumn{2}{c}{Instance CD (valid)} & Validity \\
\cmidrule(lr){2-7} \cmidrule(lr){8-10} \cmidrule(lr){11-12} \cmidrule(lr){13-13}
& ACC$_{cmd}\uparrow$ & ACC$_{param}\uparrow$ & Line$\uparrow$ & Arc$\uparrow$ & Circle$\uparrow$ & Ext.$\uparrow$ & COV$\uparrow$ & MMD$\downarrow$ & JSD$\downarrow$ & Mean$\downarrow$ & Median$\downarrow$ & IR$\downarrow$ \\
\midrule
DeepCAD & 62.3 & 54.1 & 66.0 & 46.8 & 49.5 & 59.2 & 72.4 & 2.05 & 4.88 & 0.95 & 0.64 & 55.2 \\
Hand2CAD & 78.4 & 69.7 & 82.0 & 61.6 & 65.3 & 72.9 & 80.7 & 1.72 & 4.12 & 0.75 & 0.49 & 48.7 \\
Drawing2CAD-Raster & 79.8 & 73.1 & \rev{84.6} & \rev{63.2} & \rev{69.1} & \rev{78.4} & \rev{82.3} & \rev{1.62} & \rev{3.99} & 0.72 & \rev{0.45} & 43.5 \\
OrthoLatentCAD & \textbf{84.2} & \textbf{76.0} & \textbf{87.5} & \textbf{70.4} & \textbf{73.1} & \textbf{80.6} & \textbf{85.8} & \textbf{1.48} & \textbf{3.76} & \textbf{0.62} & \textbf{0.40} & \textbf{39.8} \\
\bottomrule
\end{tabular}}
\caption{Full metrics for the 1,000 real participant--part three-view drawing sets. Results use F+R+T inputs without real-set fine-tuning. ACC and IR include all examples; geometry metrics use compilable predictions. ACC, COV, and IR are percentages; MMD and JSD are scaled by $10^2$, and CD is scaled by $10^3$.}
\label{tab:supp_real_full}
\end{table*}

\section{Responsible Use and Deployment Considerations}
OrthoLatentCAD is intended to provide an editable first-pass CAD initialization from orthographic drawing references, particularly during early mechanical and product-design exploration.
It is not an autonomous replacement for expert CAD modeling: kernel compilation establishes executability, but does not guarantee dimensional accuracy, design intent, manufacturability, or fitness for use.

The principal deployment risk is therefore over-reliance on a plausible or compilable output.
Before downstream simulation, manufacturing, or safety-critical use, a designer should verify the recovered dimensions and constraints, inspect the construction history, rerun CAD-kernel checks, and correct any geometric or procedural errors.

\section{\texorpdfstring{\rev{View Availability and Boundary Case}}{View Availability and Boundary Case}}
\rev{This section examines the effect of view availability on reconstruction in a real-hand boundary case.}

\begin{figure*}[tbp]
\centering
\includegraphics[width=\linewidth]{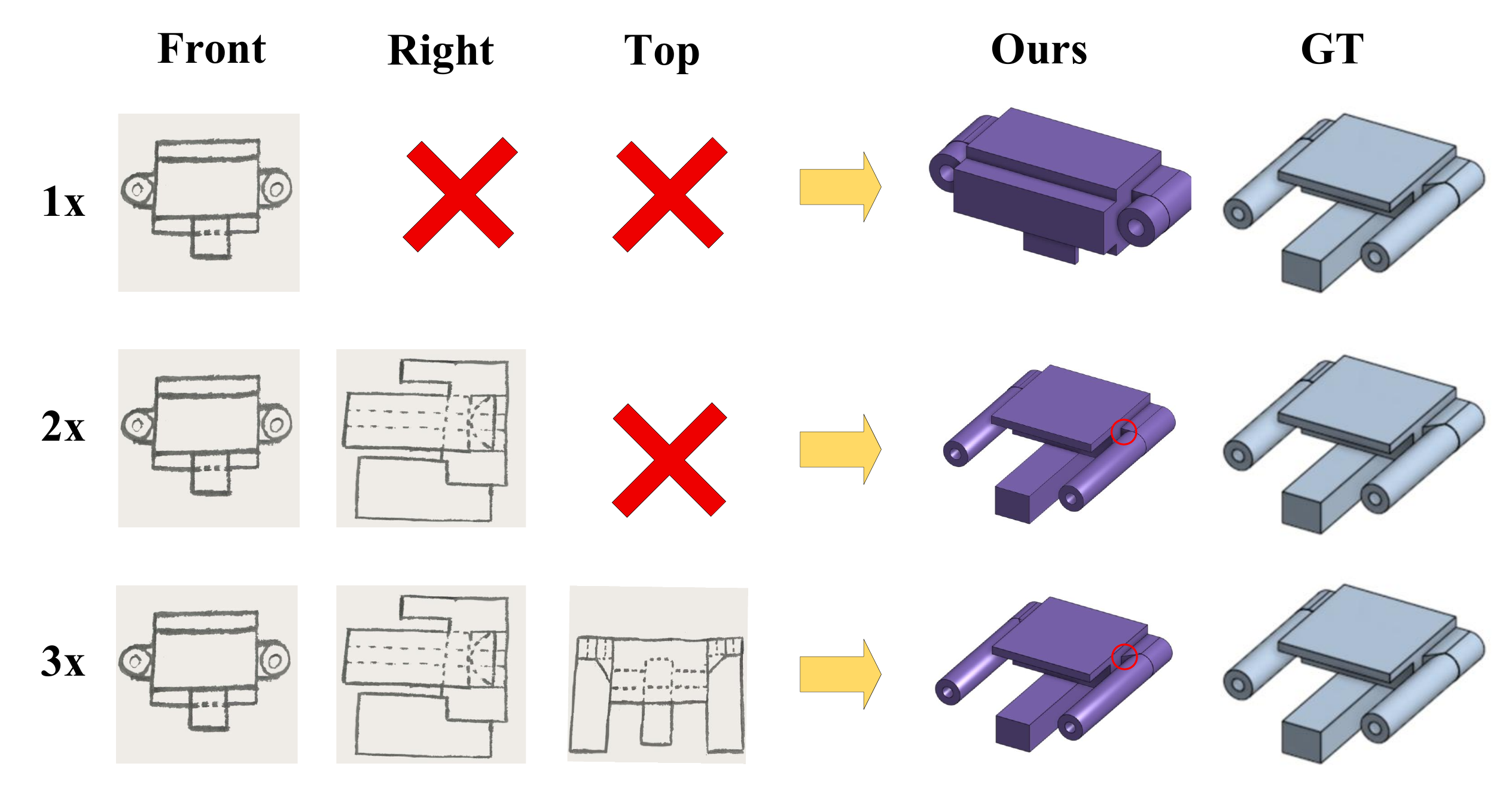}
\caption{Effect of view availability in a real-hand boundary case. The 1x, 2x, and 3x rows use F, F+R, and F+R+T, respectively. Additional views recover the global fork-and-cylinder structure, but the red circles mark a local cylinder--body junction that remains inaccurate.}
\label{fig:supp_boundary_cases}
\end{figure*}

Figure~\ref{fig:supp_boundary_cases} separates improvement from residual error as views are added.
The front view alone underdetermines the part; adding the right view recovers most of its elongated structure, and the top view further constrains the layout.
Nevertheless, the local cylinder--body junction remains inaccurate, showing that additional projections reduce ambiguity but do not enforce exact sketch connectivity or dimensional constraints.
This failure is consistent with the numerical loop and local-constraint limitations discussed in the main paper and motivates constraint-aware decoding and kernel-guided repair.

\clearpage
{\small
\bibliographystyle{ieeenat_fullname}
\bibliography{main}
}